\documentclass{article}

\usepackage[final]{corl_2020} 

\usepackage{times}
\usepackage{algorithmic}
\usepackage{algorithm}
\usepackage{amsmath}
\usepackage{amssymb}
\usepackage{graphics}
\usepackage{epsfig}
\usepackage[caption=false, font=footnotesize]{subfig}
\usepackage{multirow}
\usepackage{wrapfig}
\usepackage{amsthm}
\usepackage{mathtools}
\usepackage{makecell}
\usepackage{bbm}

\DeclareMathOperator*{\argmax}{arg\,max}

\title{Policy learning in $SE(3)$ action spaces}

\author{
    Dian Wang \quad \quad \quad 
    Colin Kohler \quad \quad \quad 
    Robert Platt\\
    Khoury College of Computer Sciences\\
    Northeastern University\\
    United States\\
    \texttt{\{wang.dian, kohler.c, r.platt\}@northeastern.edu}
}

\begin{document}
\maketitle

\theoremstyle{definition}
\newtheorem{definition}{Definition}


\begin{abstract}

In the spatial action representation, the action space spans the space of target poses for robot motion commands, i.e. $SE(2)$ or $SE(3)$. This approach has been used to solve challenging robotic manipulation problems and shows promise. However, the method is often limited to a three dimensional action space and short horizon tasks. This paper proposes ASRSE3, a new method for handling higher dimensional spatial action spaces that transforms an original MDP with high dimensional action space into a new MDP with reduced action space and augmented state space. We also propose SDQfD, a variation of DQfD designed for large action spaces. ASRSE3 and SDQfD are evaluated in the context of a set of challenging block construction tasks. We show that both methods outperform standard baselines and can be used in practice on real robotics systems.

\end{abstract}



\section{Introduction}
Many applications of model free policy learning to robotic manipulation involve an action space where actions correspond to small displacements of the robot joints or gripper position. While this approach is simple and flexible, it often necessitates reasoning over long time horizons (hundreds of time steps) in order to solve even moderately complex manipulation tasks, e.g. to build the block structure shown in Figure~\ref{fig:envs}. An alternative is the spatial action representation where actions correspond to destination positions and orientations in the workspace~\cite{zeng2018roboticpick-and-place,pushinggrasping,platt_aaai2019}. For each valid destination pose, the corresponding action executes an end-to-end arm motion that reaches that pose. Spatial action representations in $SE(2)$ have been shown to dramatically accelerate policy learning for both manipulation and navigation applications~\cite{spatial_navigation,platt_aaai2019}. However, this method is generally limited to three dimensional action spaces and low complexity tasks.

This paper makes two contributions that can help scale up the spatial action representation approach to handle more challenging tasks. First, we propose a $Q$ function representation that can span up to six dimensions of pose in $SE(3)$ rather than just the three dimensions that span $SE(2)$. The method is based on an augmented state space approach where an MDP with a large action space is converted into a new equivalent MDP with a smaller action space but a larger state space~\cite{rl_binary_search,upper_lower_mdp}. The second contribution is a novel imitation learning method (a modified version of DQfD~\cite{dqfd}) that is better suited to the large action spaces inherent to spatial action representations than standard baseline methods. Imitation learning is important in our setting because it can guide exploration in complex sparse reward tasks where unstructured exploration would be very unlikely to sample a goal state. We evaluate our algorithms in the context of challenging block construction tasks that require the agent to learn how to stably grasp and place objects, how to sequence placements in order to achieve the desired structure, and how to handle novel object heights and complex object shapes during construction. We show that the proposed methods outperform standard baselines on these tasks and that the resulting policies can be run directly on a real robotic system.

\section{Related Work}

\noindent
\textbf{Dense affordance prediction:} In dense affordance prediction, the agent makes predictions about the one-step outcome of an action, e.g. grasping, pushing, insertion, etc. Early applications of this idea were in grasp detection where a fully convolutional network (FCN) was used to encode grasp quality for pick points at each pixel in a scene~\cite{mahler2017dex,zeng2018roboticpick-and-place}. The idea was extended to encoding action values for dense affordance prediction problems where the agent predicts the one-step outcomes of pushing for grasping~\cite{pushinggrasping, pushing_grasping_reward,ShiftingObjectsforGrasping}, the outcomes of a shovel-and-grasp motion primitive~\cite{Slide-to-Wall}, robotic tossing~\citep{tossingbot}, and kit assembly~\citep{form2fit}. In contrast, this paper uses a dense affordance-like model to encode $Q$ values in a reinforcement learning setting where the agent selects actions based on their return over a long time horizon. As a result, our agent can reason about construction of multi-step structures rather than just the outcome of one or two actions.

\noindent
\textbf{Spatial action representations in DQN:} In the spatial action representation, the action space of a robot is encoded as a set of destination poses in $SE(2)$. This representation was used by~\cite{nair2018overcoming} and~\cite{li2019towards} who learn to stack block towers using per-block reward feedback in a three DOF ($x,y,z$) subset of $SE(3)$. \citet{platt_aaai2019} learns policies that can stack four-block towers using sparse rewards and an action space that spans three DOF ($x,y,\theta$) of $SE(2)$. \citet{gualtieri2018learning} learns manipulation policies in action spaces that span six DOF of $SE(3)$. However, actions are sampled sparsely using a grasp detector rather than densely. More recently, \citet{spatial_navigation} demonstrate DQN using a three DOF action representation ($x,y,\theta$) where the $Q$ function is encoded using an FCN. In contrast to the above, our paper learns policies expressed over a densely sampled 6-DOF action space of $SE(3)$ rather than the 3-DOF action space of $SE(2)$. This gives our agent flexibility to reason about out of plane orientations that are usually absent in model free approaches to manipulation.

\noindent
\textbf{Robotic Block Stacking:} Learning block construction policies can be very challenging. Work in this area often assumes dense rewards, decompose the problem into a sequence of simpler one-block stack policies, or focus on perception at the expense of control. For example, \citet{deisenroth2011learning} use PILCO to learn a parameterized short-horizon policy that stacks a single block on top of a tower. This policy is iteratively executed to construct larger stacks. \citet{nair2018overcoming} and \citet{li2019towards} leverage an imitation learning method for learning block stacking policies that assume the agent is provided with the exact positions of all blocks and dense sequential rewards for stacking each successive block. \citet{janner2018reasoning} focuses on learning object-oriented representations from visual data but ignores the problem of robotic control nearly completely. Similarly, \citet{groth2018shapestacks} predict the stability of a stack based on images as input, but ignores the control problem. Our paper goes beyond the work above in two key ways: 1) we learn policies over images rather than object positions; 2) our agent is only rewarded for achieving a final desired structure. We are unaware of other work that can produce similarly complex structures under these conditions.

\noindent
\textbf{Factored action spaces:} Another strand of related work is the idea of factoring the action space into a set of loosely related variables. For example, \citet{sallans2004reinforcement} factors large state/action spaces into loosely related subproblems. \citet{factor_atari} factors the Atari action space into horizontal movement, vertical movement, and fire. \citet{action_branching} also learns $Q$ values for different action dimensions separately. These approaches are more efficient than the augmented state approach, but in contrast to this paper, they make strong conditional independence assumptions about the $Q$ values of the different action dimensions.

\noindent
\textbf{The augmented state approach to large action spaces:} A key idea of our paper is to reduce the size of the action space by increasing the size of the state space. We call this the \textit{augmented state approach}. An early instantiation of this idea was in the context of linear programming approaches to MDPs by~\cite{de2004constraint} who proposed converting the selection of a single high dimensional action into a sequence of partial action selections. \citet{pazis2011reinforcement} applied this idea in a TD learning context where they converted a single high dimensional action into a sequence of binary actions. More recently, \citet{upper_lower_mdp} proposed a ``Sequential DQN'' model that uses the same idea to select high dimensional actions in continuous control problems like HalfCheetah and Swimmer by sequentially assigning each dimension of action as a separate action. Our paper leverages the augmented state representation idea in the context of $SE(3)$ action spaces. To our knowledge, this is the first application of the augmented state approach to this setting.


\section{Problem Statement}
\label{sect:problem}

\begin{wrapfigure}[15]{R}{0.28\textwidth}
  \begin{center}
    \includegraphics[width=0.28\textwidth]{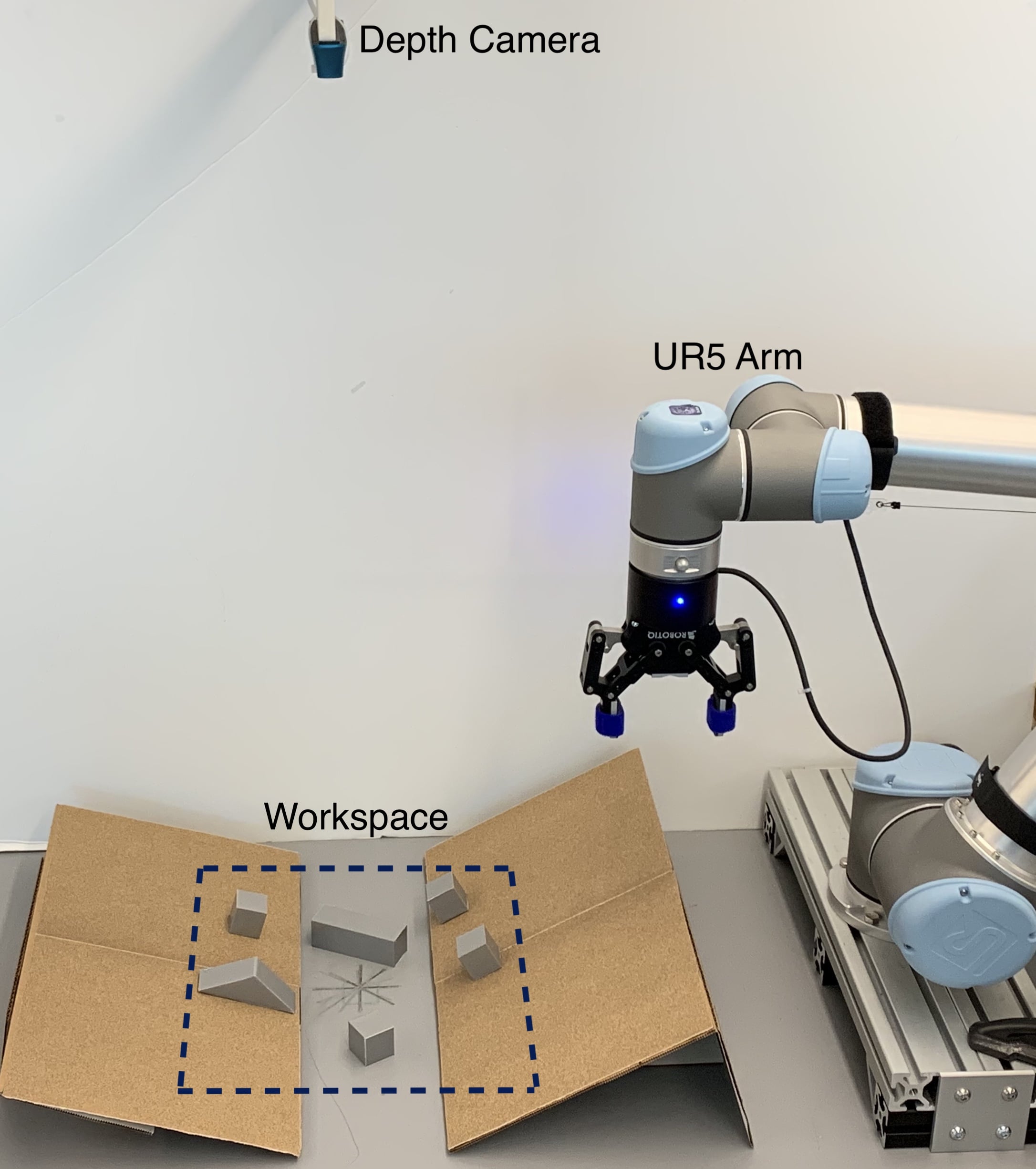}
  \end{center}
  \caption{Our manipulation setting.}
  \label{fig:workspace}
\end{wrapfigure}

We address prehensile manipulation problems where the robot's action space is defined primarily by a set of gripper destination poses, i.e. a subset of $SE(3)$. Observations are in the form of heightmap images derived from a point cloud of the manipulation scene. The scene contains objects (in our case blocks) of either known or unknown geometry that are placed arbitrarily. The objective is to learn a policy over arm and gripper motions that solves a desired manipulation assembly or construction task (see Figure~\ref{fig:workspace}). We focus on the sparse rewards setting where the agent is only rewarded for completing a desired structure or assembly.

\noindent
\textbf{Action space:} The action set $A$ is a subset of $\{ \textsc{pick}, \textsc{place} \} \times SE(3)$. Each action $a = (a^{pp}, a^{se3}) \in A$ is a collision free arm motion to $a^{se3} \in SE(3)$ (planned using an off the shelf motion planner) followed by a gripper closing action (if $a^{pp} = \textsc{pick}$) or a gripper opening action (if $a^{pp} = \textsc{place}$).

\noindent
\textbf{State space:} States are triples, $s_t = (I_t,H_t,g_t) \in S$, where $I_t$ is a top down heightmap of the scene taken at time $t$, $H_t$ is the \textit{in-hand} image, and $g_t \in \{\textsc{open}, \textsc{closed} \}$ denotes gripper state at time $t$. The in-hand image depends on the action executed on the last time step (at time $t-1$). If the last action was a $\textsc{pick}$, then $H_t$ is a set of heightmaps that describe the 3D volume centered and aligned with the gripper frame $a^{se3}_{t-1} \in SE(3)$ when the \textsc{pick} occurred. Otherwise, $H_t$ is set to the zero value image.

\noindent
\textbf{Reward:} We use a sparse reward function that is one for all state action pairs that reach the goal state and zero otherwise.

\section{Approach}


\subsection{The augmented state representation}

\label{sect:augmented_state}

\begin{wrapfigure}[8]{r}{0.4\textwidth}
\vspace{-0.5cm}
  \begin{center}
    \includegraphics[width=0.4\textwidth]{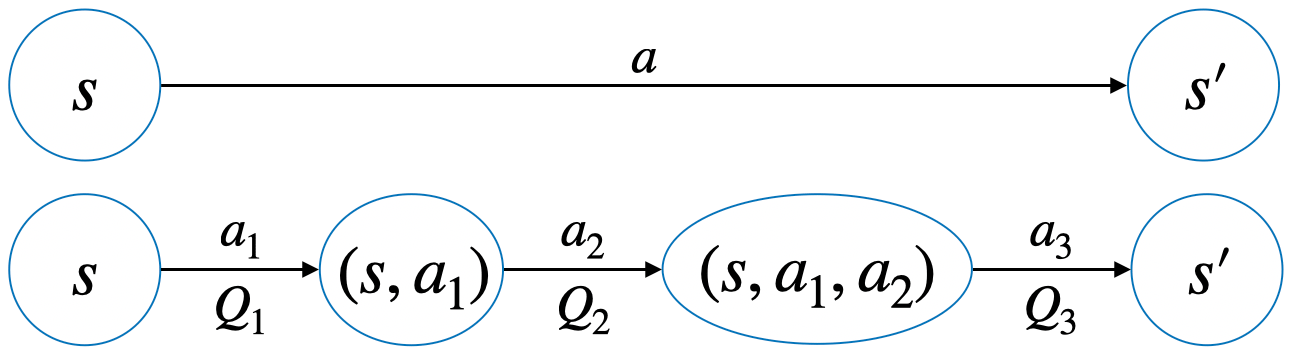}
  \end{center}
  \caption{Augmented state representation for $A = A_1 \times A_2 \times A_3$.}
  \label{fig:augmented_mdp_transform}
\end{wrapfigure}

The augmented state approach~\cite{rl_binary_search,upper_lower_mdp} transforms a given MDP $\mathcal{M}$ into a new MDP $\bar{\mathcal{M}}$ with more states but fewer actions. In particular, suppose $\mathcal{M}$ has state space $S$ and action space $A = A_1 \times \dots \times A_n$. We will refer to an action variable $A_i$ as a ``partial action''. We define a new MDP $\bar{\mathcal{M}}$ where each single action $a \in A$ in $\mathcal{M}$ corresponds to a sequence of $n$ partial actions $a_1 \in A_1, \dots, a_n \in A_n$ in $\bar{\mathcal{M}}$. The state space of $\bar{\mathcal{M}}$ is augmented with additional states that ``remember'' the sequence of partial actions performed so far since the last transition in $\mathcal{M}$. 
Specifically, the new MDP $\bar{\mathcal{M}}$ has an action space $\bar{A} = A_1 \cup \dots \cup A_n$ and a state space $\bar{S} = S_1 \cup \dots \cup S_n$ where $S_1 = S$ and $S_i = S \times A_1 \times \dots \times A_{i-1}$ for $i\geq2$. After executing $a_1$ from state $s$, $\bar{\mathcal{M}}$, transitions deterministically to $(s,a_1)$, and to $(s,a_1,a_2)$ after executing $a_2$ after that, and so on. This is illustrated in Figure~\ref{fig:augmented_mdp_transform}. The top of Figure~\ref{fig:augmented_mdp_transform} shows a transition in $\mathcal{M}$ and the bottom shows the same transition in $\bar{\mathcal{M}}$. The two intermediate states $(s_1,a_1)$ and $(s_1,a_1,a_2)$ ``remember'' the first two partial actions while the third partial action causes a transition to $s'$ that mirrors the original transition in $\mathcal{M}$. The transitions to the intermediate states have 0 rewards $\bar{r}=0$ and a discount factor $\bar\gamma=1$, while the transition from $(s, a_1, a_2)$ to $s'$ has the same reward and discount factor as the original MDP.


\subsection{ASRSE3: Augmented state representation in $SE(3)$ action spaces}


\begin{wrapfigure}[22]{R}{0.55\textwidth}
\centering
    \includegraphics[width=0.55\textwidth]{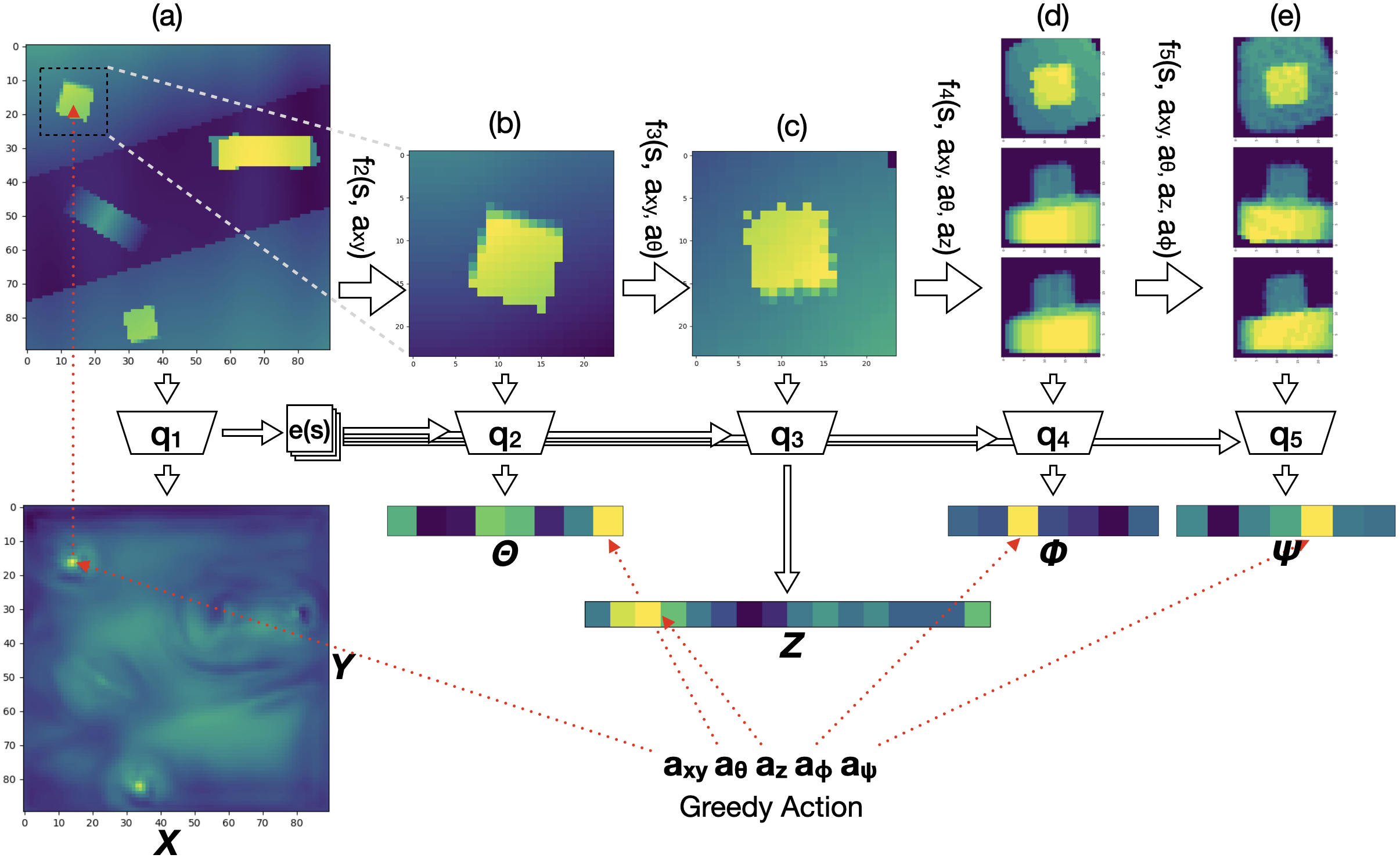}
    \vspace{-0.5cm}
    \caption{Our model selects actions $a_1=a_{xy}$, $a_2=a_\theta$, $a_3=a_{z}$, $a_4=a_\phi$, and $a_5=a_\psi$ sequentially. $q_1$ maps the heightmap image onto a $|X| \times |Y|$ map of $Q$ values with a maximum at $a_{xy}$. Given that selection of $a_{xy}$, $q_2$ predicts $Q$ values for $a_\theta$. Similarly, $q_3$, $q_4$, and $q_5$ predict $Q$ values for $a_{z}$, $a_\phi$, and $a_\psi$, respectively. The input of in-hand image $H$ and the output head selection accomplished by $g$ is omitted in the figure. (a): the top-down heightmap $I$. (b-e) the output of $f_2$-$f_5$.}
    \label{fig:pipeline}
\end{wrapfigure}

We focus on problems with action spaces that span six DOF of $SE(3)$: $a^{se3} = (x,y,z,\theta,\phi,\psi)$, where $x,y,z \in X \times Y \times Z$ denote position and $\theta,\phi,\psi \in \Theta \times \Phi \times \Psi$ denote orientation in terms of ZYX Euler angles. We apply the augmented state approach by setting the augmented action space to be $\bar{A} = A_1 \cup \dots \cup A_5$ with $A_1 = X \times Y$, $A_2 = \Theta$, $A_3 = Z$, $A_4 = \Phi$, and $A_5 = \Psi$, and setting the augmented state space to be $\bar{S} = S_1 \cup \dots \cup S_5$ where $S_1 = S$ and $S_i = S \times A_1 \times \dots \times A_{i-1}$ for $2\leq i\leq 5$. The question becomes how to encode a $Q$ function over the the augmented state and action space. Recall that states are triples, $s = (I,H,g) \in S$, where $I$ is a top-down heightmap of the scene, $H$ is an in-hand image that describes the contents of the hand, and $g$ denotes grasp status.

We define a series of $Q$ functions $Q_i : S_i \times A_i \rightarrow \mathbb{R}$ for $i \in \{1, \dots, 5\}$. $Q_1$ is encoded using a fully convolutional U-Net model, $q_1(s) \rightarrow \mathbb{R}^{|X| \times |Y|}$ that maps each $(x,y)$ pixel in the scene image $I$ to a $Q$ value, c.f.~\cite{zeng2018roboticpick-and-place,pushinggrasping}. $Q_2$ is encoded by $q_2$, a small convolutional network that takes as input: 1) $e(s)$, a feature encoding of the scene image generated by $q_1$; 2) the in-hand image $H$; 3) the output of a function $f_2(s,a_1) \rightarrow \mathbb{R}^{m \times m}$ that generates an $m \times m$ crop of $I$ centered at $a_1=(x,y)$ (Figure~\ref{fig:pipeline}b). The purpose of $f$ is to encode the selected partial actions into images. $q_2$ outputs a vector in $\mathbb{R}^{|\Theta|}$. $Q_3$ is encoded by $q_3$ which is similar to $q_2$ except that it uses a cropping function $f_3(s,a_1,a_2) \rightarrow \mathbb{R}^{m \times m}$ which applies the $f_2$ transform and then rotates the output by $a_2 = \theta$ (Figure~\ref{fig:pipeline}c). $q_3$ outputs a vector in $\mathbb{R}^{|Z|}$. $q_4$ and $q_5$ ($Q_4$ and $Q_5$) are similar to $q_3$ except that $f_3(s,a_1,a_2)$ is replaced with $f_4(s,a_1,a_2,a_3)$ and $f_5(s,a_1,a_2,a_3,a_4)$ which apply successive additional dimensions $a_3 = z$ and $a_4 = \phi$. Unlike $f_2$ and $f_3$, $f_4$ and $f_5$ output three orthographic projections ($\mathbb{R}^{3\times m\times m}$) of the cropped region, viewing it along the rotated $z$, $y$, and $x$ axes (Figure~\ref{fig:pipeline}d, e, see Appendix~\ref{appen:state_action_enc} for details). $q_4$ and $q_5$ output $\mathbb{R}^{|\Phi|}$ and $\mathbb{R}^{|\Psi|}$ respectively. All five networks have two output heads for \textsc{pick} and \textsc{place}, $g$ controls which output head to use. The full neural network model is given in Appendix~\ref{appen:network}.

\noindent
\textbf{Training the model:} Each of the five $Q$ functions can be trained using standard DQN methods, i.e. by minimizing a loss $\mathcal{L}_i = \mathbb{E} \big(y_i - Q_i(\bar{s},\bar{a}) \big)^2$ for a TD target $y_i$. The standard 1-step TD loss would yield $y_1 = \max_{a_2} Q_2((s,a_1),a_2)$, $y_2 = \max_{a_3} Q_3((s,a_1,a_2),a_3)$, $y_3 = \max_{a_4} Q_4((s,a_1, a_2, a_3), a_4)$, $y_4 = \max_{a_5} Q_5((s,a_1, a_2, a_3, a_4), a_5)$, and $y_5 = r + \gamma \max_{a'_1} Q_1(s',a'_1)$. However, we find empirically that we do better with $n$-step returns forward to the next transition in the original MDP. Specifically, we set 
$y_1 = y_2 = y_3 = y_4 = y_5 = r + \gamma\max_{a'_5} Q_5((s',a'^*_1,a'^*_2, a'^*_3, a'^*_4),a'_5)$ where $a'^*_1=\argmax_{a'_1}Q_1(s', a'_1)$, $a'^*_2=\argmax_{a'_2}Q_2((s', a'^*_1), a'_2)$, $a'^*_3=\argmax_{a'_3}Q_3((s', a'^*_1, a'^*_2), a'_3)$, $a'^*_4 = \argmax_{a'_4} Q_4((s',a'^*_1,a'^*_2, a'^*_3), a'_4)$
A comparison between 1-step and $n$-step returns is in Appendix~\ref{appen:n-step}. 

Henceforth, we will refer to this approach as ASRSE3 (Augmented State Representation for $SE(3)$).

\subsection{Imitation learning in large action spaces}
\label{sect:sdqfd}

The complexity of the tasks that we want to learn makes it hard to use standard model free algorithms without additional supervision during learning. Here, we guide exploration using imitation learning. Both DQfD~\cite{dqfd} and ADET~\cite{adet} have proven to be effective in domains like Atari. These methods learn $Q$ values while penalizing the values of non-expert actions. DQfD incorporates a finite penalty into its targets for non-expert actions using a large margin loss, $\mathcal{L}_{LM} = \mathbb{E}_{(s,a^e)} \big[ \max_a [Q(s,a) + l(a^e, a)]-Q(s, a^e) \big]$, where $a^e$ denotes the expert action from state $s$. ADET accomplishes something similar using a cross entropy term which tends toward negative infinity as the learned policy departs from the expert policy. ADET can learn faster than DQfD in large action spaces because the cross entropy term penalizes $Q$ values for \emph{all} non-expert actions in a given state -- not just for the one with the highest $Q$ value. However, the fact that the cross entropy loss tends toward negative infinity for non-expert actions can cause serious errors in the resulting $Q$ estimates. Is it possible to get the best of both worlds?

We propose a method that we call \emph{Strict} DQfD (SDQfD). The key idea is to apply the large margin loss penalty to all feasible actions that have a $Q$ value larger than the expert action minus the non-expert action penalty. Let $A^{s,a^e}$ denote the set of actions for which this is the case: $A^{s,a^e} = \big\{ a \in A \big| Q(s,a) > Q(s, a^e)-l(a^e, a) \big\}$. 

Now, we define the ``strict'' large margin loss term:
\begin{equation}
\mathcal{L}_{SLM} = \frac{1}{|A^{s,a^e}|}\sum_{a \in A^{s,a^e}} \Big[Q(s, a) + l(a^e, a) - Q(s, a^e) \Big]
\label{eqn:slm}
\end{equation}
In contrast to DQfD which applies the large margin penalty to only a single non-expert action, Equation~\ref{eqn:slm} applies a penalty to all actions $a \in A^{s,a^e}$ that are within a margin of the value of the expert action. As we show later, this method significantly outperforms DQfD in large action spaces.

\noindent
\textbf{Generating expert trajectories:} In order to use imitation learning, we need to obtain expert trajectories. While human expertise is often called upon to provide this guidance, we use regression planning (\citet{stuart2016artificial}, Ch 11) to generate successful goal-reaching trajectories by searching backward from a goal state. Starting with a stable goal state block structure, we code an expert deconstruction policy that simply picks up the highest block and places it on the ground with a random position and orientation. By reversing the deconstruction episode, we can acquire an expert construction episode.

\noindent
\textbf{SDQfD with ASRSE3:} Note that we can immediately apply SDQfD (or other imitation learning algorithms) in the ASRSE3 framework because the augmented state MDP is still an MDP. We simply replace the the $n$-step TD loss described in Section~\ref{sect:augmented_state} with the appropriate imitation learning loss. When ASRSE3 is combined with SDQfD, DQfD, DQN, etc., we call the resulting algorithm ASRSE3 SDQfD, ASRSE3 DQfD, ASRSE3 DQN, etc.


\section{Experiments}
\label{sect:exps}

\begin{figure}[t]
\newlength{\env}
\setlength{\env}{0.09\linewidth}
\centering
\subfloat[2S]{\includegraphics[width=\env]{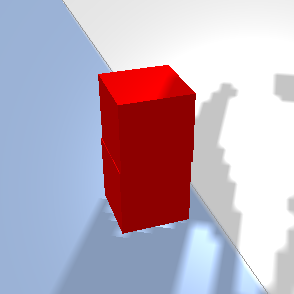}\label{fig:envs_2s}}
\subfloat[4S]{\includegraphics[width=\env]{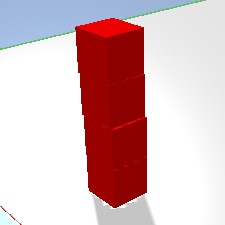}\label{fig:envs_4s}}
\subfloat[4H1]{\includegraphics[width=\env]{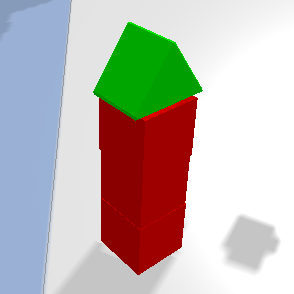}\label{fig:envs_4h1}}
\subfloat[5H1]{\includegraphics[width=\env]{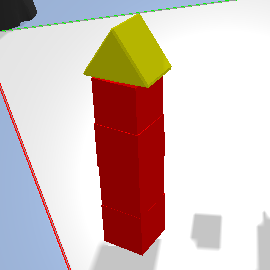}\label{fig:envs_5h1}}
\subfloat[H2]{\includegraphics[width=\env]{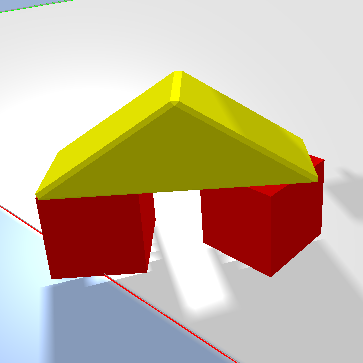}\label{fig:envs_h2}}
\subfloat[H3]{\includegraphics[width=\env]{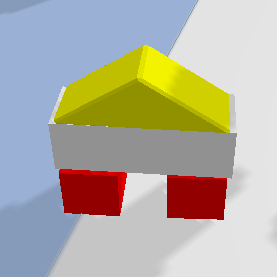}\label{fig:envs_h3}}
\subfloat[H4]{\includegraphics[width=\env]{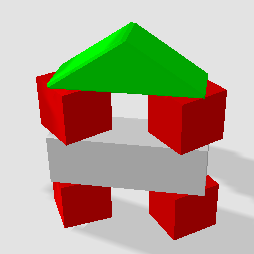}\label{fig:envs_h4}}
\subfloat[ImDis]{\includegraphics[width=\env]{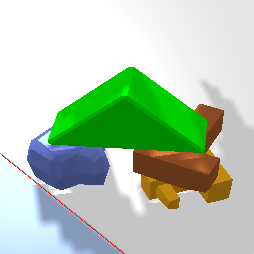}\label{fig:envs_imdis}}
\subfloat[ImRan]{\includegraphics[width=\env]{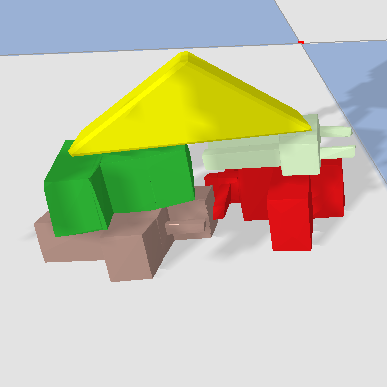}\label{fig:envs_imran}}
\subfloat[ImH2]{\includegraphics[width=\env]{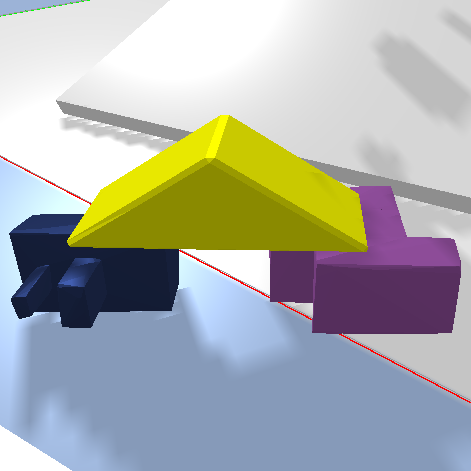}\label{fig:envs_imh2}}
\subfloat[ImH3]{\includegraphics[width=\env]{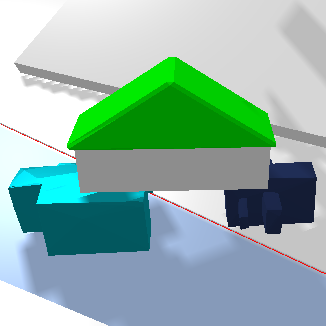}\label{fig:envs_imh3}}
\caption{Experimental environments.}
\label{fig:envs}
\end{figure}

We evaluate ASRSE3 and SDQfD both separately and together in the context of the 11 block construction tasks shown in Figure~\ref{fig:envs}. In all cases, the robot is presented with the requisite blocks of known or novel geometry placed at random poses in the environment. The robot must grasp each block and place it stably in order to achieve the desired structure as shown. In tasks (a) through (g), the blocks are the regular known shapes as shown. In tasks (h) through (k), the block heights are randomized and the shapes are sampled from a large set of challenging shapes (see Appendix~\ref{appen:task}). In Section~\ref{sect:exp_dqn}-\ref{sect:exp_6d}, both training and testing are in PyBullet simulation~\cite{pybullet}. Section~\ref{sec:exp_robot} tests the trained model in the real world both in the known and novel block settings.

\subsection{ASRSE3 DQN}
\label{sect:exp_dqn}

This experiment compares ASRSE3 DQN 
with a baseline version of DQN and DDPG on the following three block stacking tasks on a flat workspace: 2S, 4S, H2 (see Figure~\ref{fig:envs}). We perform this evaluation only in an $SE(2)$ action space (i.e. $x,y,\theta$) in order to accommodate the baselines which cannot work in more complex environments. The $z$ coordinate is controlled heuristically given $a_{xy}$ by evaluating the depth of the height map at $a_{xy}$. Our baseline DQN approach (FCN DQN) encodes the $Q$ function using a similar strategy as used by~\citet{spatial_navigation,pushinggrasping}. We use the same U-Net architecture we use in our $Q_1$ network in ASRSE3. For a given state $s=(I,H,g)$, it outputs a feature map where each pixel encodes the $Q$ value of the action corresponding $x,y$ coordinate. We handle the $\theta$ dimension by inputting a stack of scene images $I_\theta$ for eight discrete values of $\theta$. In DDPG, the backbone used in the actor and critic is a ResNet-34 architecture. The action is represented as a 3-vector $x,y,\theta$. In ASRSE3, we factor the action space into two variables, $\bar{A} = A_1 \cup A_2$ with $A_1=X\times Y, A_2=\Theta$, and use two $Q$ functions, $Q_1(s,a_1)$ and $Q_2((s,a_1),a_2)$. In order to make these learning tasks as simple as possible, we disallow all pick and place actions that do not move the gripper above an existing object in the scene.


\begin{wrapfigure}[12]{r}{0.6\textwidth}
\vspace{-0.7cm}
  \begin{center}
    \subfloat[2S]{\includegraphics[width=0.2\textwidth]{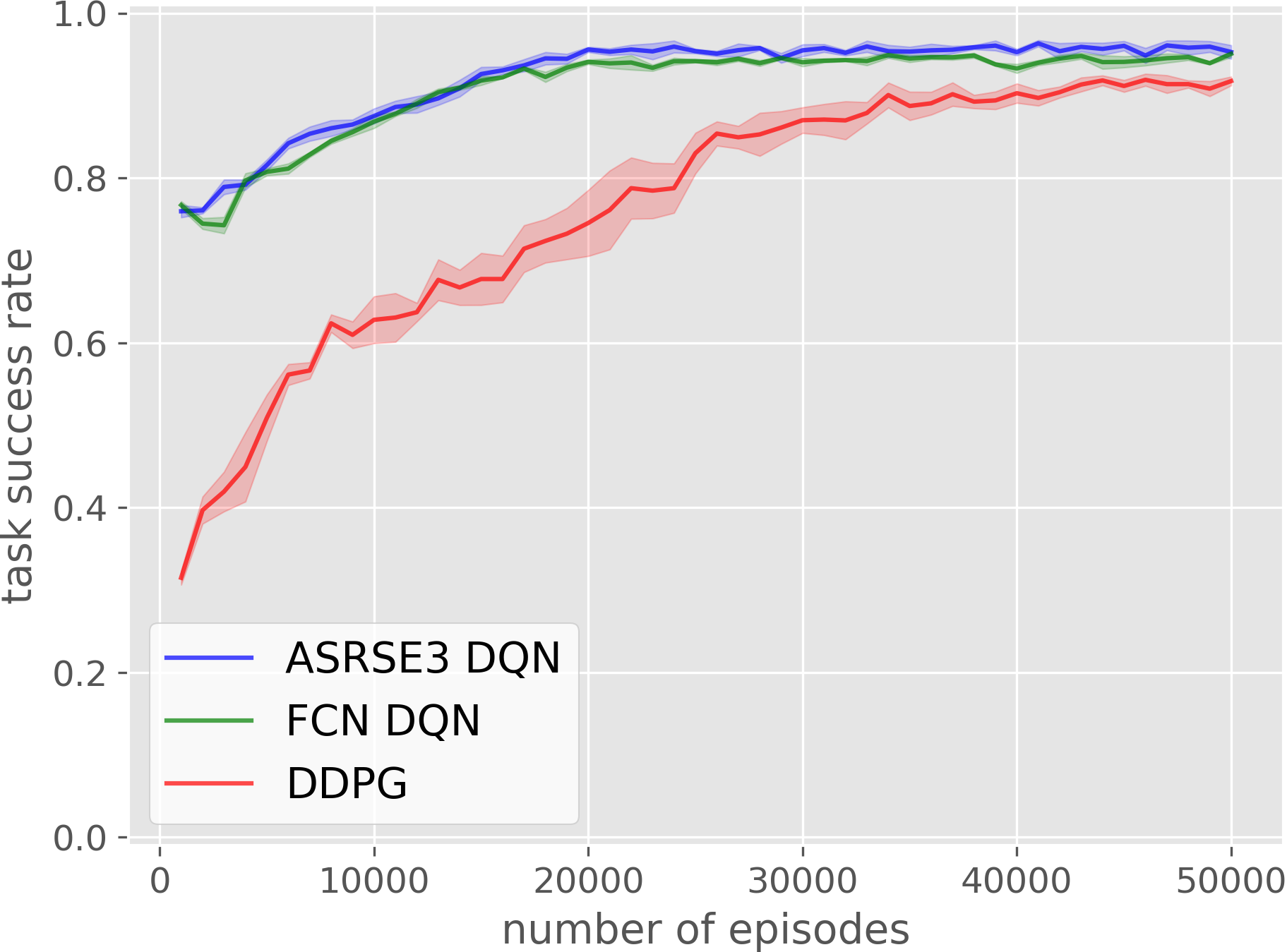}}
    \subfloat[H2]{\includegraphics[width=0.2\textwidth]{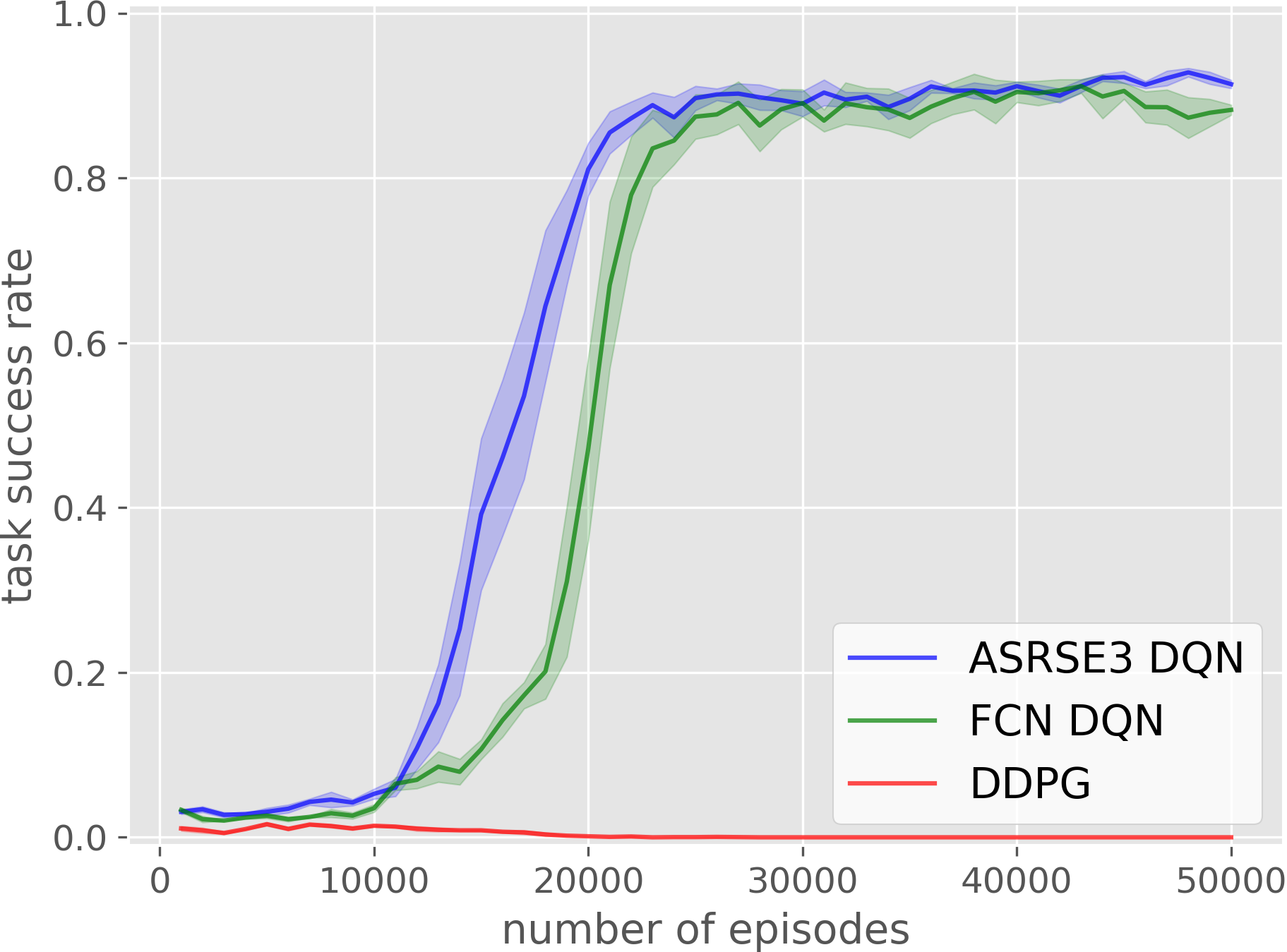}}
    \subfloat[4S]{\includegraphics[width=0.2\textwidth]{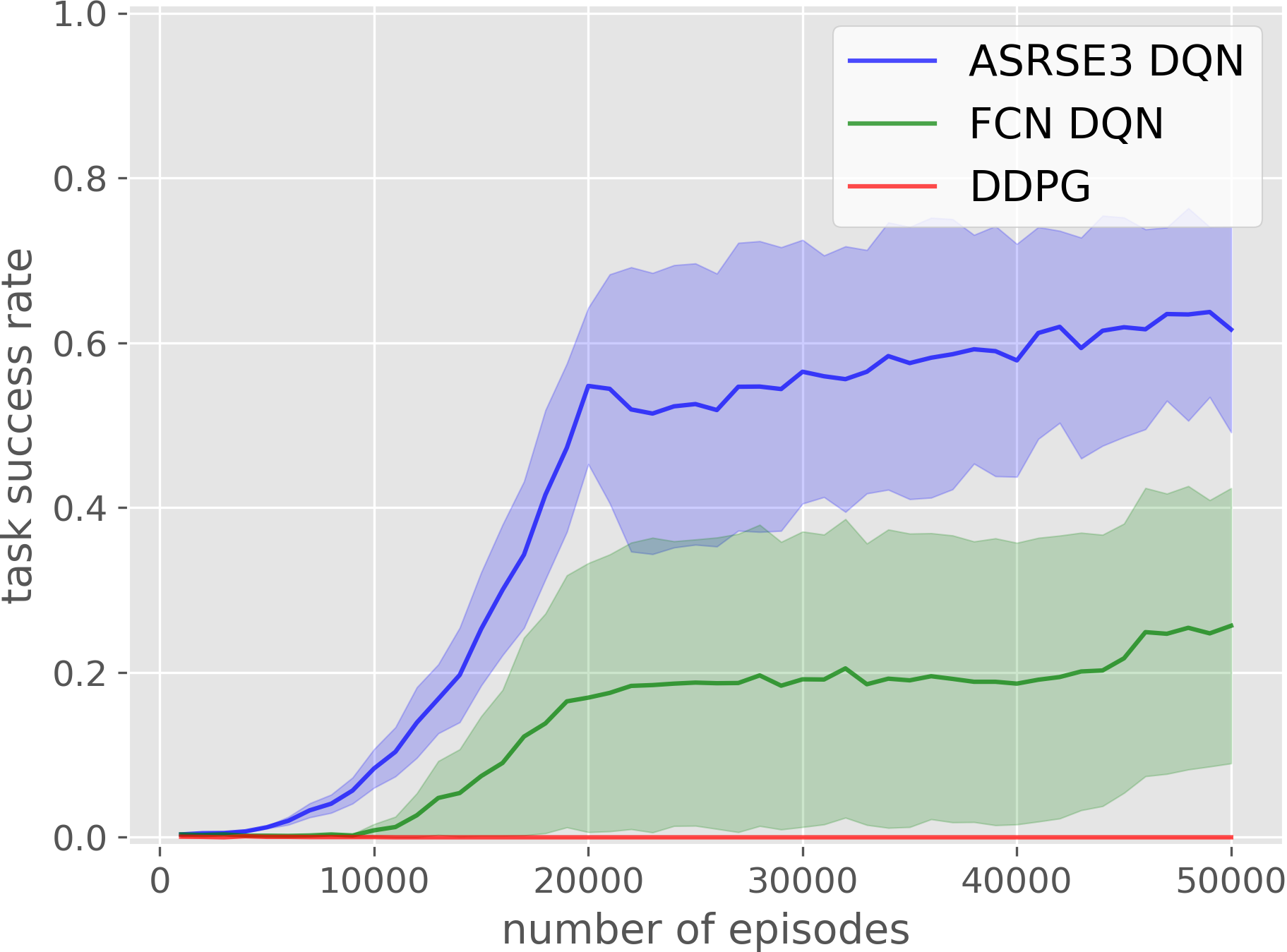}}
  \end{center}
  \vspace{-0.4cm}
  \caption{Comparison between DQN with ASRSE3 (blue), a baseline version of DQN (green), and DDPG (red). Results are averaged over 4 runs and plotted with a moving average of 1000 episodes.}
  \label{fig:exp_dqn}
\end{wrapfigure}

Figure~\ref{fig:exp_dqn} shows the results. First, notice that DDPG is only able to solve 2S, the simplest of the three tasks, and fails with the others. This is likely because representing $x, y$ as a vector deprived the model's ability to generalize over different object positions, compared to the FCN representation. Second, notice that the FCN DQN is an improvement over DDPG. It solves 2S and H2 on nearly all attempts. It is only on 4S that the FCN DQN fails on a significant fraction of attempts. This is likely because 4S is the most challenging structure to build because even a small misalignment of the topmost blocks can cause the structure to topple. In contrast, our proposed approach, ASRSE3 DQN, solves all three tasks. It solves H2 faster and it converges to a higher value for 4S.

\subsection{SDQfD versus baselines}
\label{sect:exp_sdqfd}

\begin{table}[b]
\centering
\subfloat[Average reward after 49k training episodes for each method. Best performance for each task in bold.]
{
\begin{tabular}{|@{\hskip3pt}c@{\hskip3pt}|@{\hskip3pt}c@{\hskip3pt}|@{\hskip3pt}c@{\hskip3pt}|@{\hskip3pt}c@{\hskip3pt}|@{\hskip3pt}c@{\hskip3pt}|}
\hline
 & H1 & H4 & ImDis & ImRan \\
\hline
\hline
ASRSE3 SDQfD & \textbf{0.993} & \textbf{0.930} & \textbf{0.939} & 0.763 \\
\hline
FCN SDQfD & 0.984 & \textbf{0.930} & 0.928 & \textbf{0.781} \\
\hline
\hline
ASRSE3 DQfD & 0.954 & 0.871 & 0.871 & 0.608\\
\hline
FCN DQfD & 0.699 & 0.658 & 0.803 & 0.515 \\
\hline
\hline
ASRSE3 ADET & 0.914 & 0.727 & 0.906 & 0.666\\
\hline
FCN ADET & 0.954 & 0.914 & 0.904 & 0.720\\
\hline
\end{tabular}
\label{tab:exp_hierarchy_sdqfd}
}
\hspace{0.5cm}
\subfloat[The average time per training step. Averaged over 1000 training steps.]
{
\begin{tabular}{|c|c|c|}
\hline
IL & FCN & ASRSE3 \\
Method && \\
\hline
SDQfD & 1.094s & 0.173s \\
ADET & 1.099s & 0.153s \\
DQfD & 1.084s & 0.153s \\ 
DQN & 0.357s & 0.149s \\
BC & 0.854s & 0.123s \\
\hline
\end{tabular}
\label{tab:time_exp_hierarchy_sdqfd}
}
\caption{Results of ASRSE3 SDQfD compared of baselines}
\end{table}

\begin{wrapfigure}[20]{r}{0.4\textwidth}
\vspace{-0.7cm}
\centering
\subfloat[FCN H4]{\includegraphics[width=0.2\textwidth]{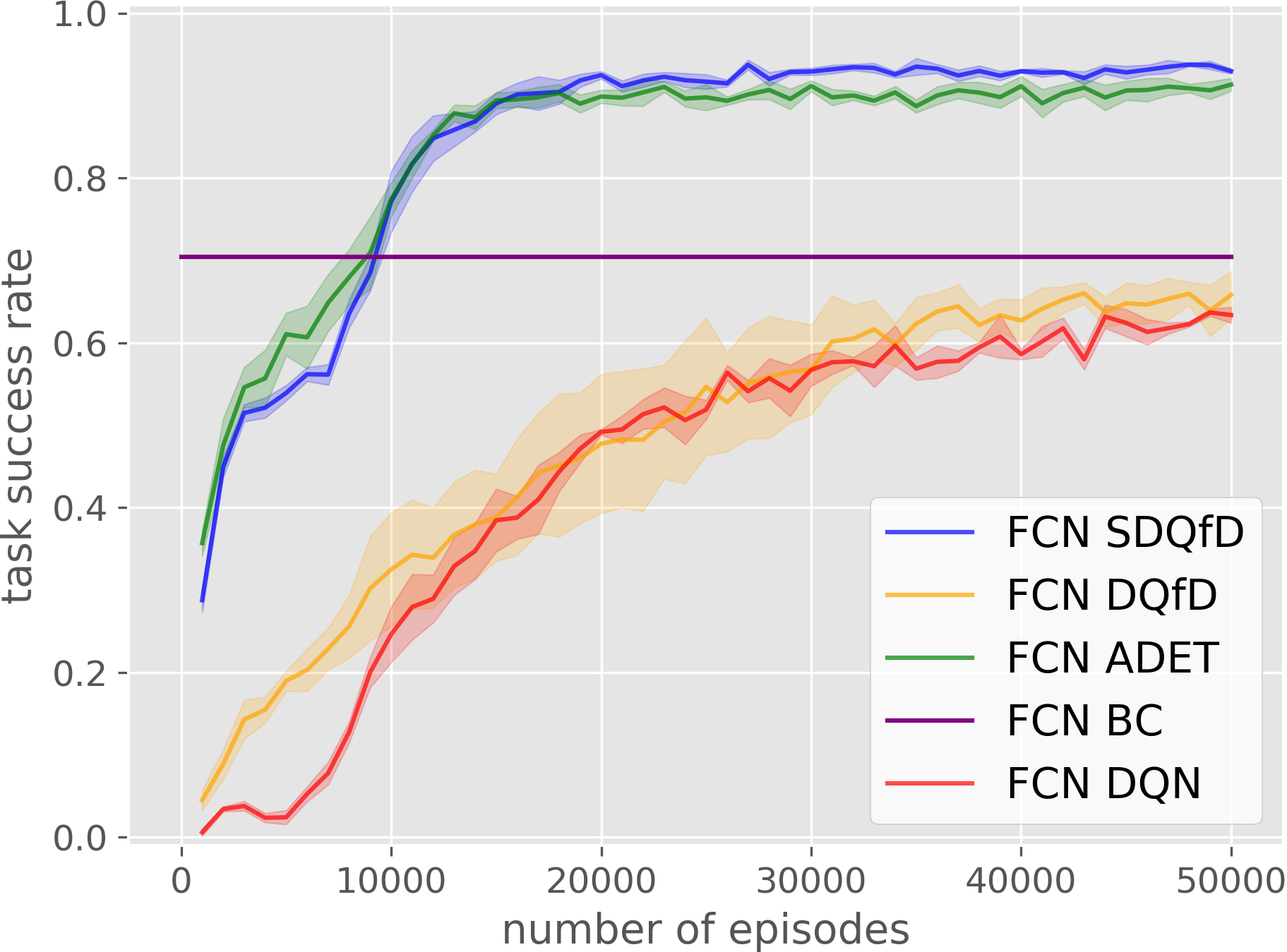}}
\subfloat[ASRSE3 H4]{\includegraphics[width=0.2\textwidth]{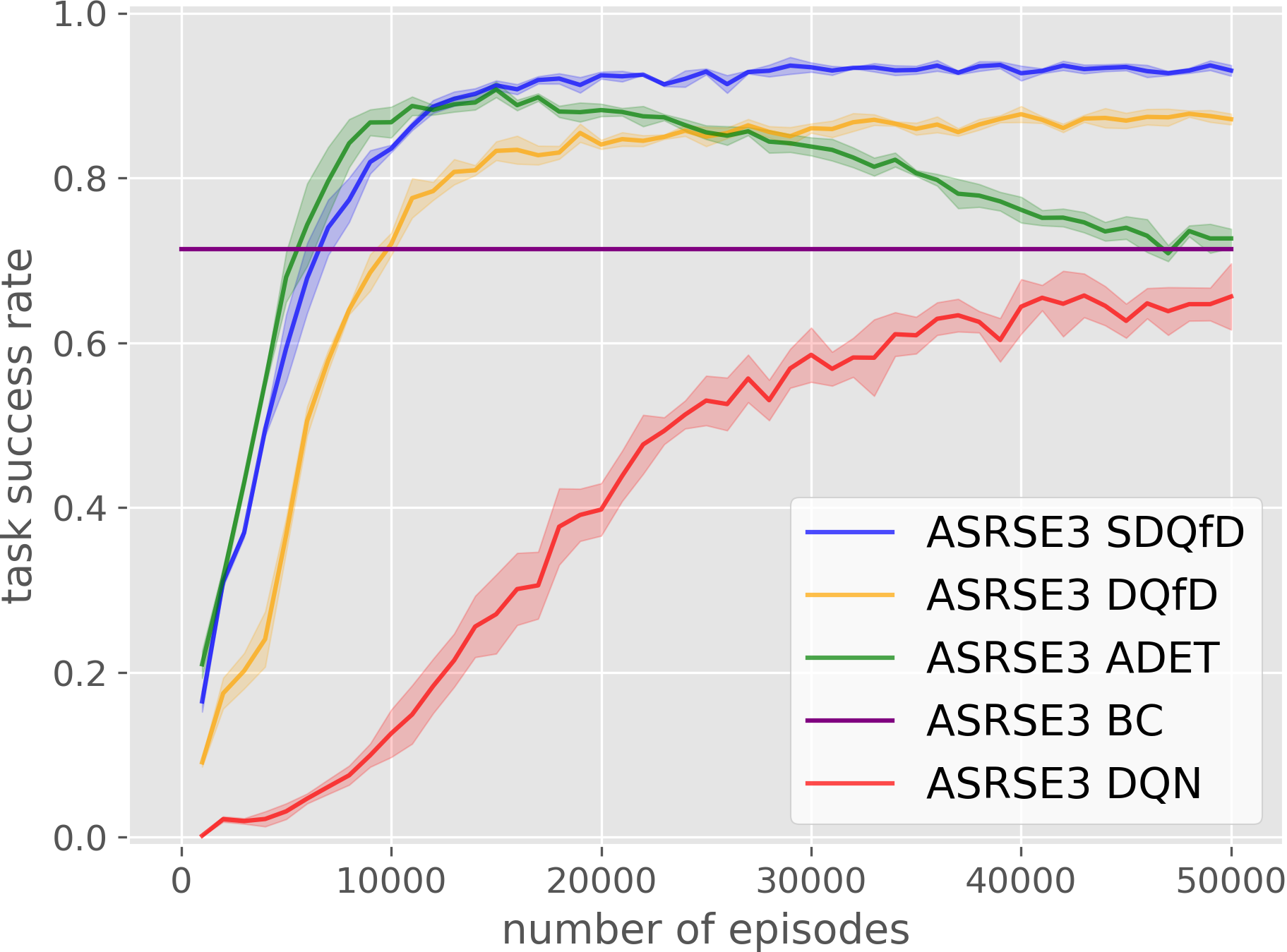}}\\
\vspace{-0.4cm}
\subfloat[FCN ImDis]{\includegraphics[width=0.2\textwidth]{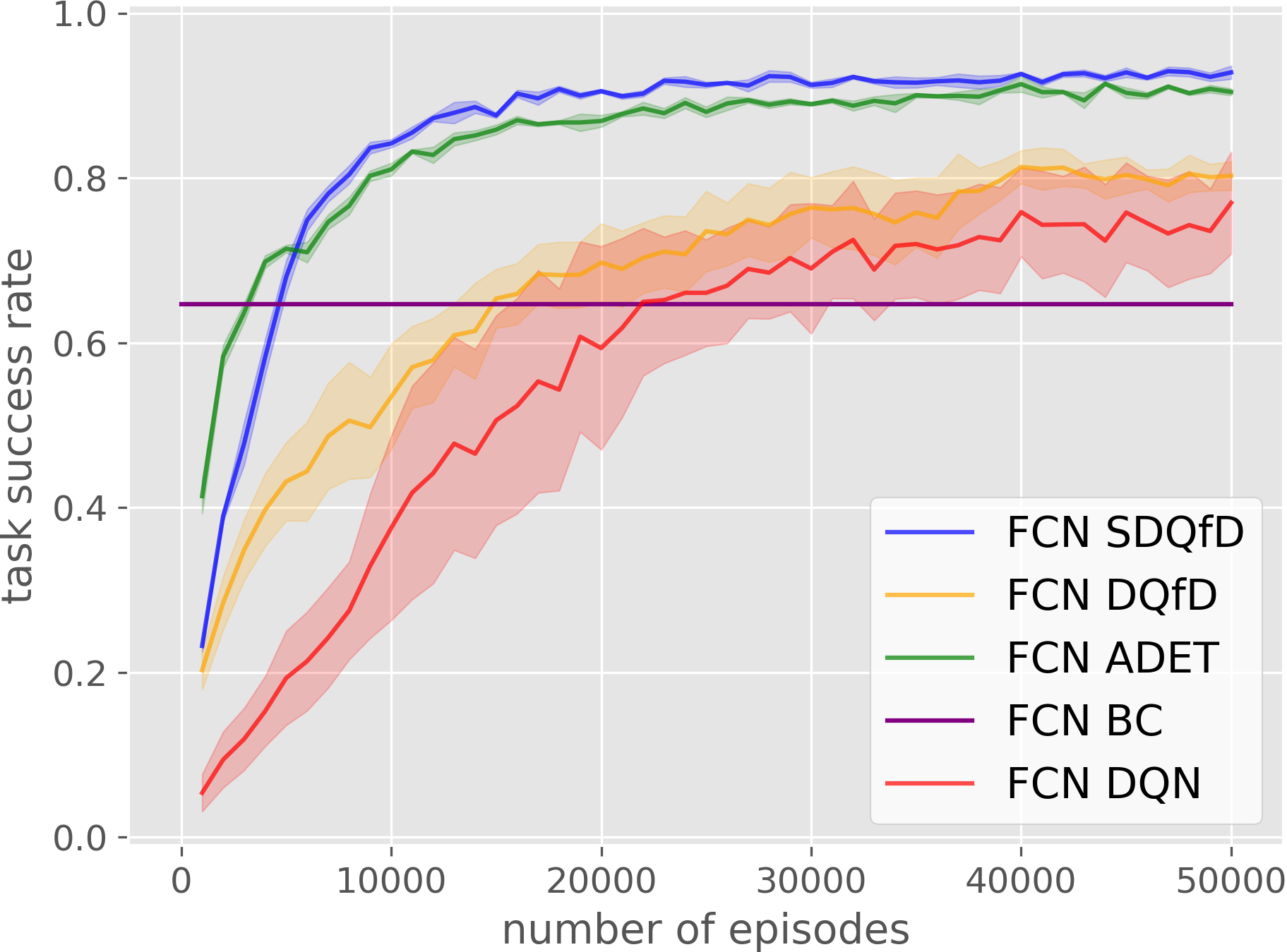}}
\subfloat[ASRSE3 ImDis]{\includegraphics[width=0.2\textwidth]{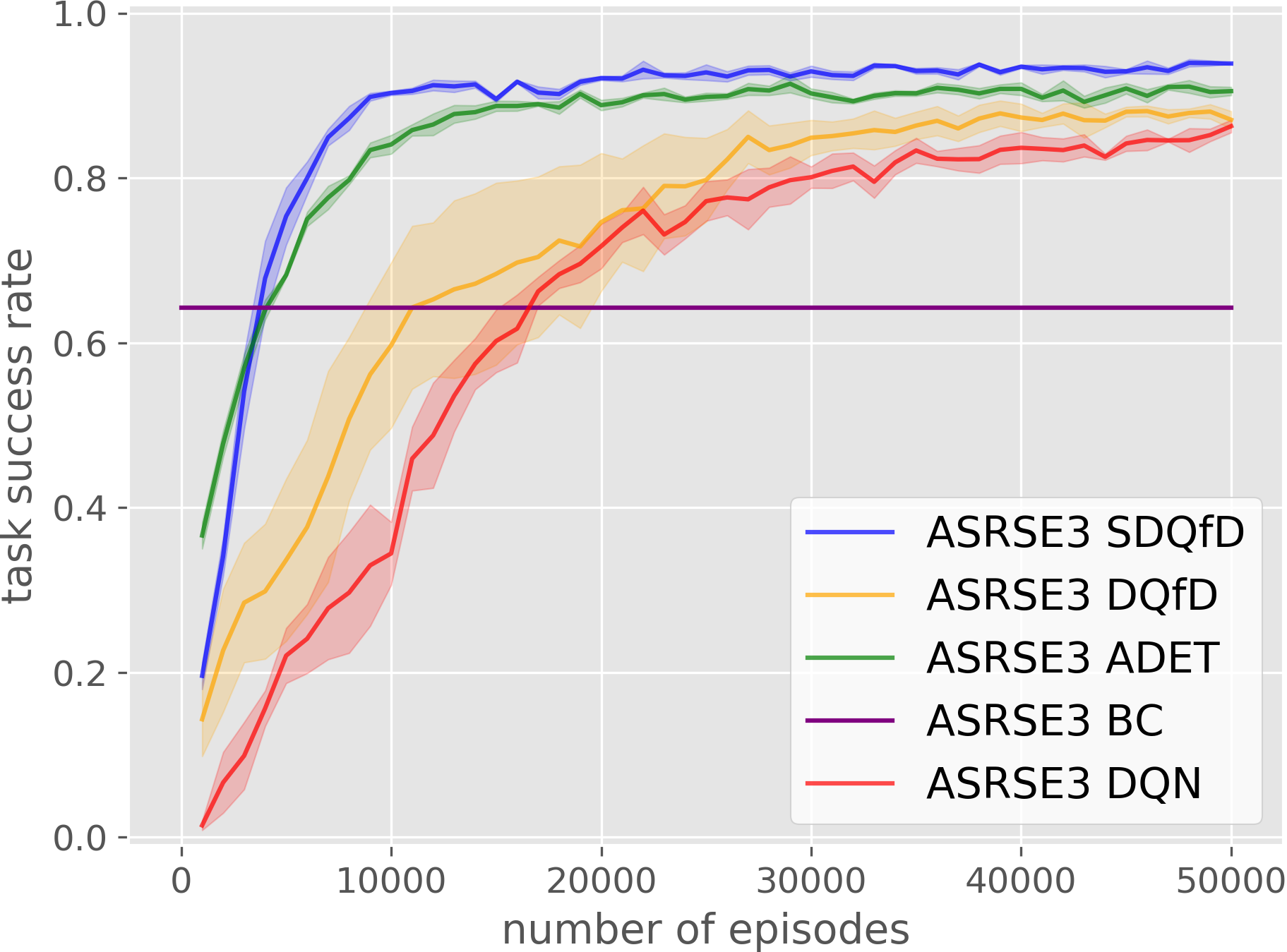}}
\vspace{-0.2cm}
\caption{(a, c) FCN SDQfD versus baseline methods. (b, d) Same comparison with ASRSE3. Results are averaged over 4 runs and plotted with a moving average of 1000 episodes.}
\label{fig:exp_sdqfd}
\end{wrapfigure}

This experiment compares SDQfD with the following baselines: 1) ADET: Accelerated DQN with expert trajectories~\cite{adet}; 2) DQfD: Deep $Q$ Learning from Demonstration~\cite{dqfd}; 3) DQN: Deep $Q$ Learning~\cite{playingatari} with pretraining; 4) BC: Naive behavior cloning~\cite{bc}. All algorithms except BC have a pretraining phase and self-play phase (see Appendix~\ref{appen:algorithm_detail}). This experiment is also performed in an $SE(2)$ action space (i.e. $x, y, \theta$). In order to evaluate SDQfD in isolation from ASRSE3, we first run SDQfD with the baseline methods using the same FCN architecture that we used in our FCN DQN baseline in Section~\ref{sect:exp_dqn}. We refer to those methods as methods with a prefix of FCN (e.g. FCN SDQfD). The results of that comparison for two challenging block construction tasks on a flat workspace, H4 and ImDis (see Figure~\ref{fig:envs}), are shown in Figure~\ref{fig:exp_sdqfd} (a, c). They show that SDQfD and ADET outperform significantly relative to either DQfD or DQN while SDQfD outperforms ADET with a small margin. Then, we compare SDQfD with the imitation learning baselines implemented using ASRSE3 in Figure~\ref{fig:exp_sdqfd} (b, d). The augmented state representation helps DQfD a lot, but SDQfD still does best. We show the same comparisons for 5H1 and ImRan tasks in Appendix~\ref{appen:sdqfd_5h1_imran}.

Table~\ref{tab:exp_hierarchy_sdqfd} compares SDQfD, DQfD, and ADET at convergence. Overall, SDQfD outperforms the best version of ADET by an average of 4\% and the best version of DQfD by an average of 16\%. Interestingly, ASRSE3 helps both SDQfD and DQfD, but it hurts ADET. We hypothesize that it huts ADET because the wildly incorrect $Q$ values induced by the cross entropy term in the ADET loss causes the incremental action selection of ASRSE3 to fail. One final point to note is that ASRSE3 is computationally much cheaper than the baseline FCN approach. Table~\ref{tab:time_exp_hierarchy_sdqfd} shows average algorithm runtimes for 1000 training steps on a single Nvidia RTX 2080Ti GPU. Although the ASRSE3 network model is larger than the baseline, it saves significant computation by not passing the $|\Theta|$ rotated version of the same image into the network.

\subsection{ASRSE3 SDQfD in a 6-DOF action space}
\label{sect:exp_6d}

\begin{figure}
\centering
\subfloat[ImH2]{\includegraphics[width=0.2\textwidth]{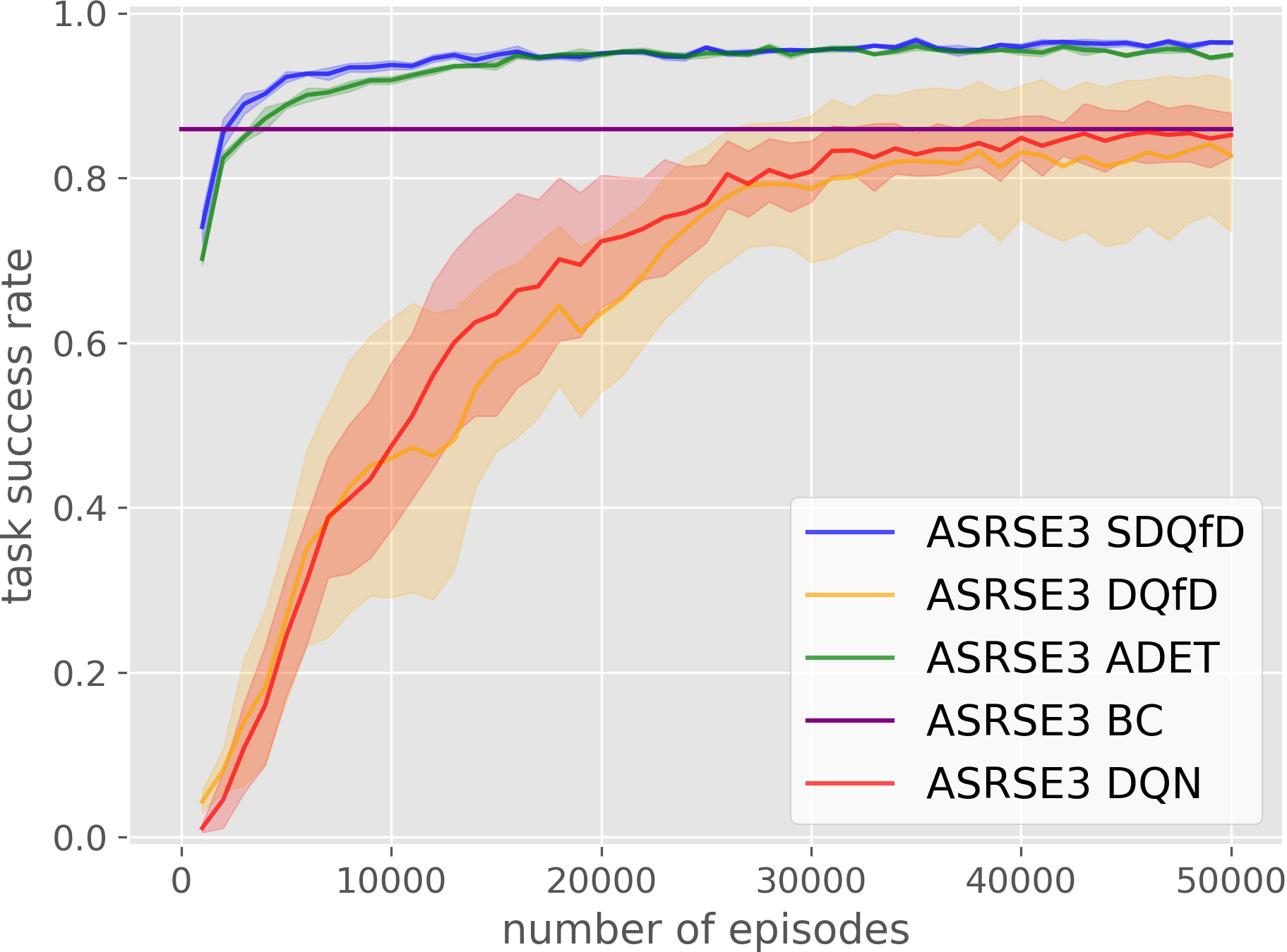}}
\subfloat[H3]{\includegraphics[width=0.2\textwidth]{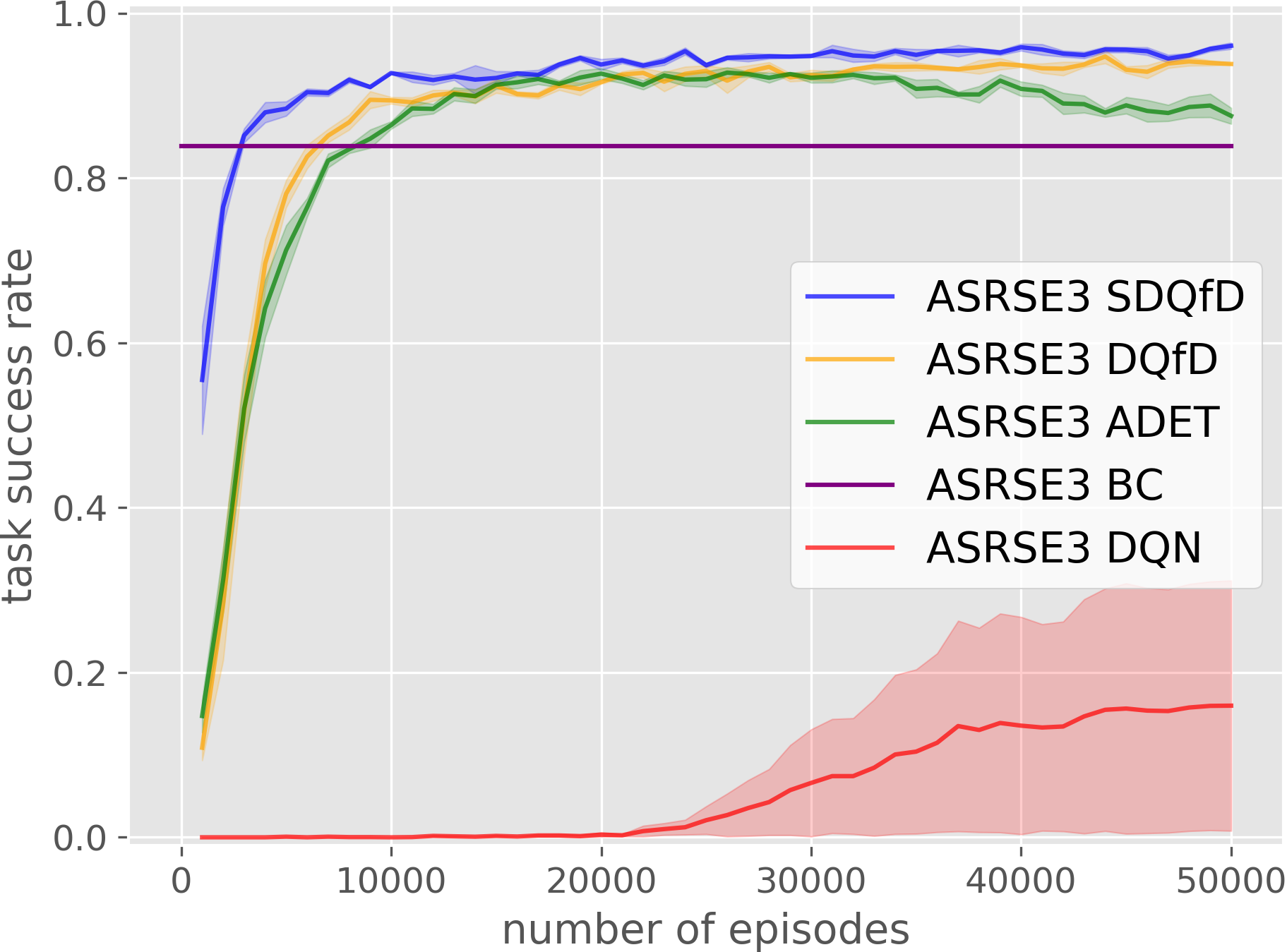}}
\subfloat[4H1]{\includegraphics[width=0.2\textwidth]{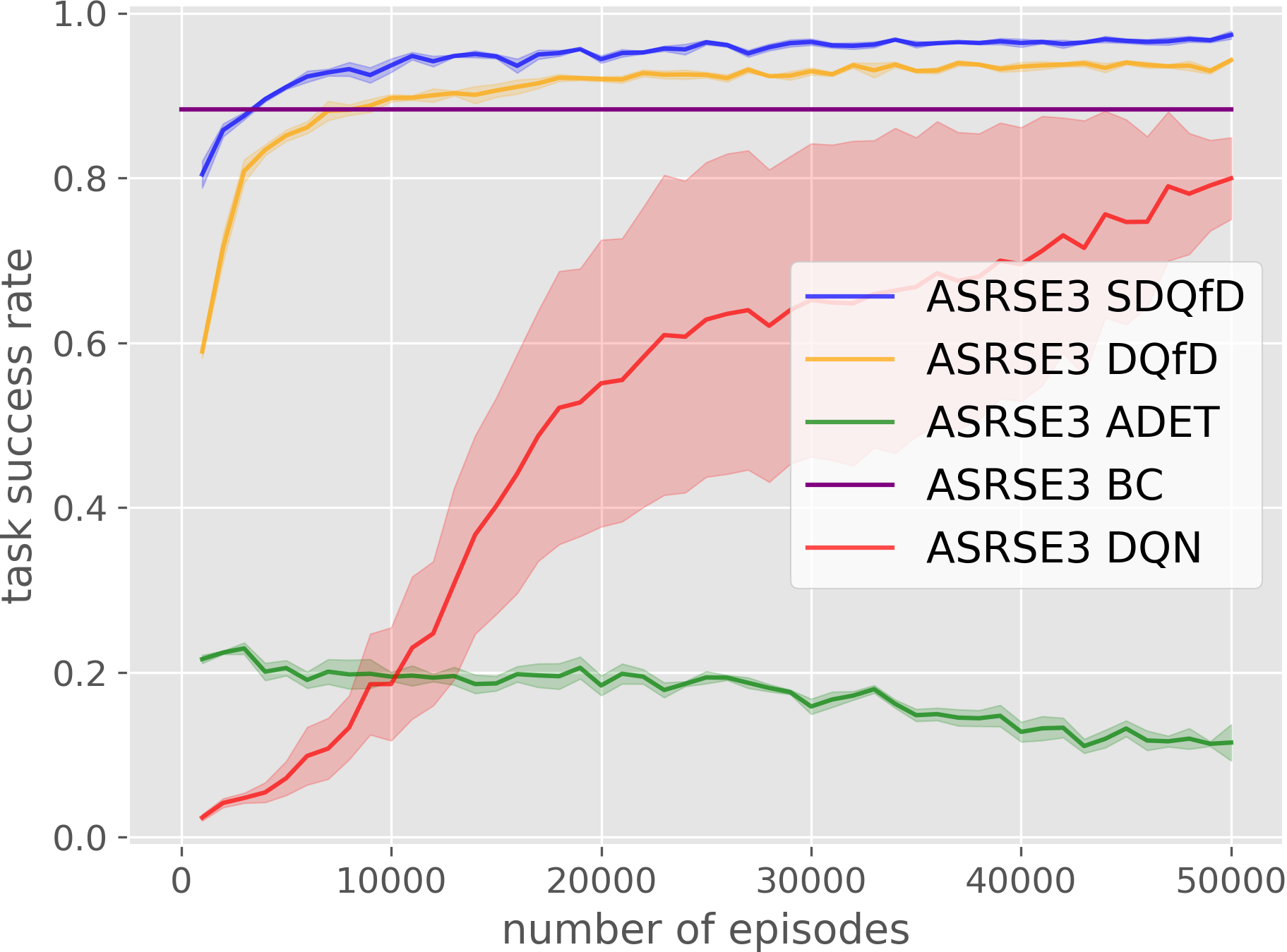}}
\subfloat[ImH3]{\includegraphics[width=0.2\textwidth]{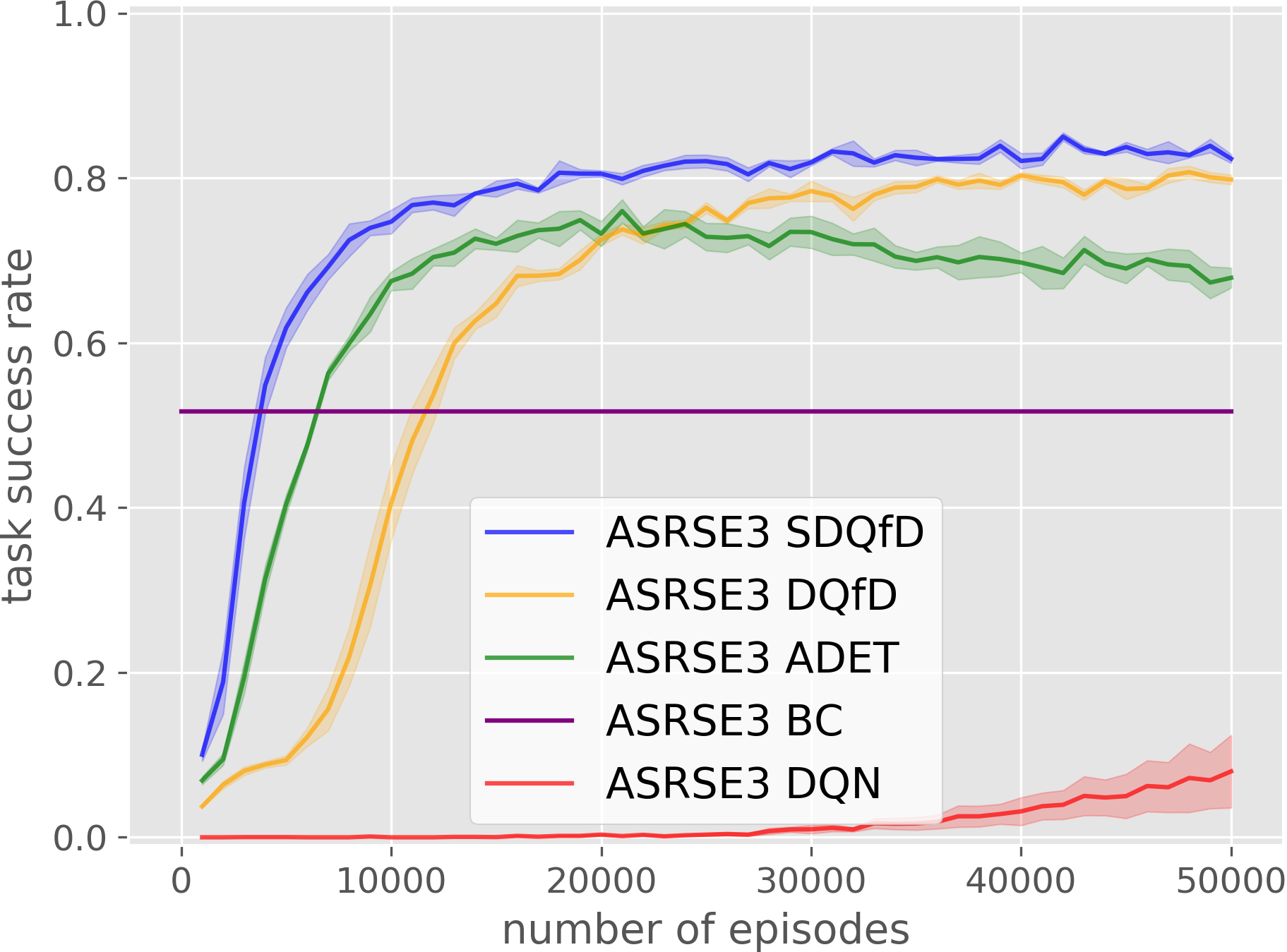}}
\subfloat[H4]{\includegraphics[width=0.2\textwidth]{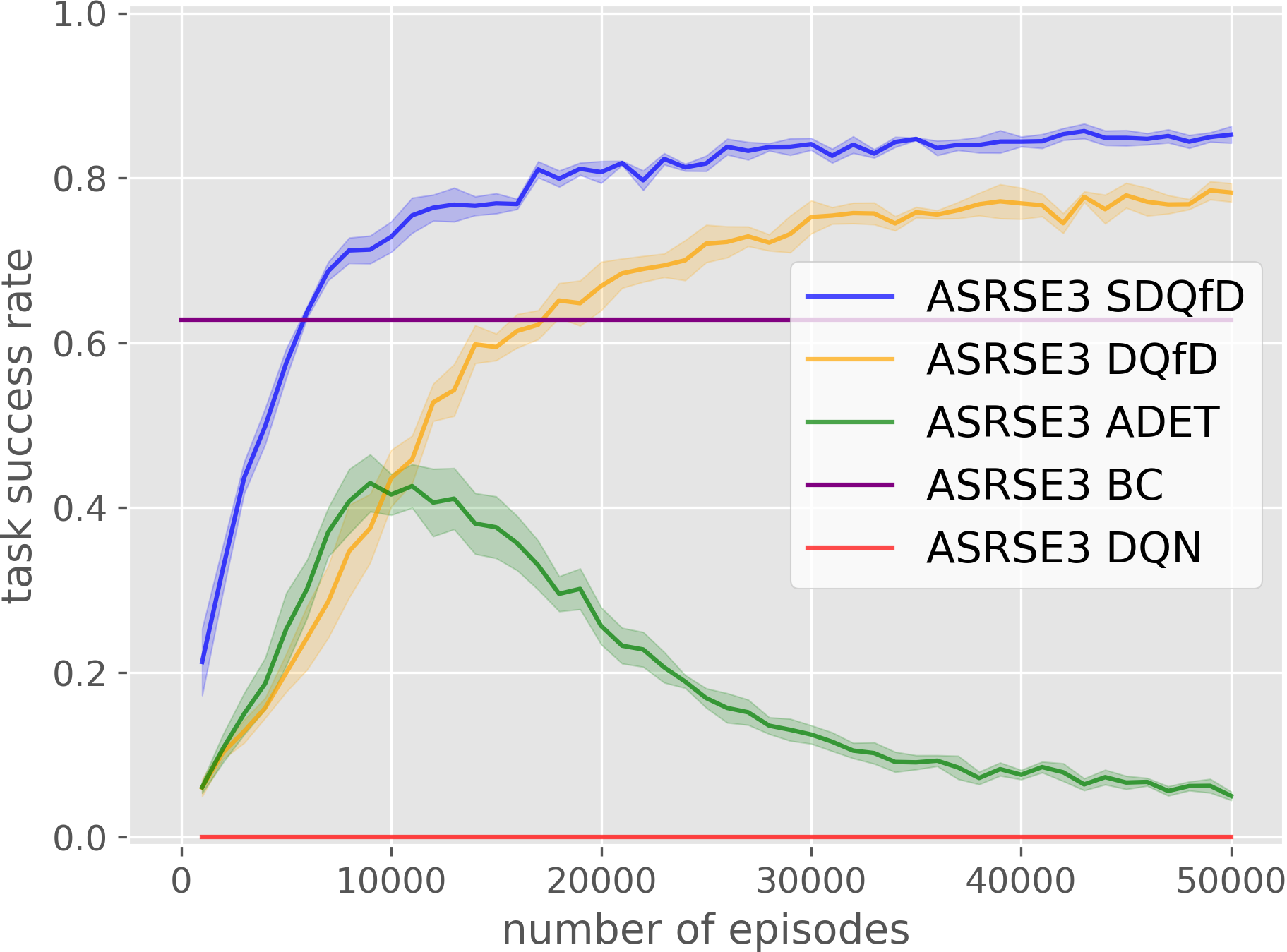}}
\caption{SDQfD versus baselines for ImH2, H3, 4H1, ImH3, and H4 tasks using the ASRSE3 $Q$ representation in the ramp domain. Results are averaged over 4 runs and plotted with a moving average of 1000 episodes. The FCN baseline is infeasible in the 6-DOF action space.}
\label{fig:exp_slope}
\end{figure}

In Sections~\ref{sect:exp_dqn} and~\ref{sect:exp_sdqfd}, the agent operates in a three dimensional action space, $x,y,\theta$. In this section, we evaluate the ability of our method to reason over six dimensions of $SE(3)$: $x,y,z,\theta,\phi,\psi$, where $\theta$, $\phi$, and $\psi$ are ZYX Euler angles.
To stimulate the agent to actively control $\phi$ and $\psi$, we created a new block construction domain involving the two ramps shown in Figure~\ref{fig:workspace}. The two ramps are always parallel to each other and the following ramp parameters are sampled uniformly randomly at the beginning of each episode: distance between ramps between 4cm and 20cm, orientation of the two ramps between 0 and 180deg, slope of each ramp between 0 and 30deg, height of each ramp above the ground between 0cm and 1cm. In addition, the relevant blocks are initialized with random positions and orientations either on the ramps or on the ground. 

We compare SDQfD, ADET, and DQfD on the ImH2, H3, 4H1, ImH3, and H4 structure building tasks using the ASRSE3 $Q$ value representation (Figure~\ref{fig:exp_slope}). Note that it is infeasible to run the FCN baseline in this setting because the action space is six DOF rather than three DOF. In all five domains, ASRSE3 SDQfD outperforms all baselines. Also, note that the amount by which SDQfD outperforms DQfD and ADET grows as the domains become more challenging.

\subsection{Robot Experiments}
\label{sec:exp_robot}

\begin{figure}[t]
\newlength{\robot}
\setlength{\robot}{0.13\linewidth}
\centering
\subfloat[]{
\includegraphics[width=\robot]{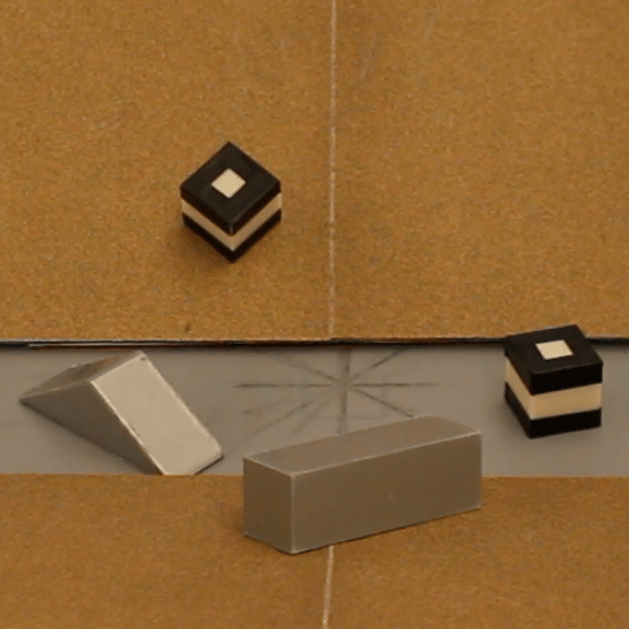}}
\subfloat[]{
\includegraphics[width=\robot]{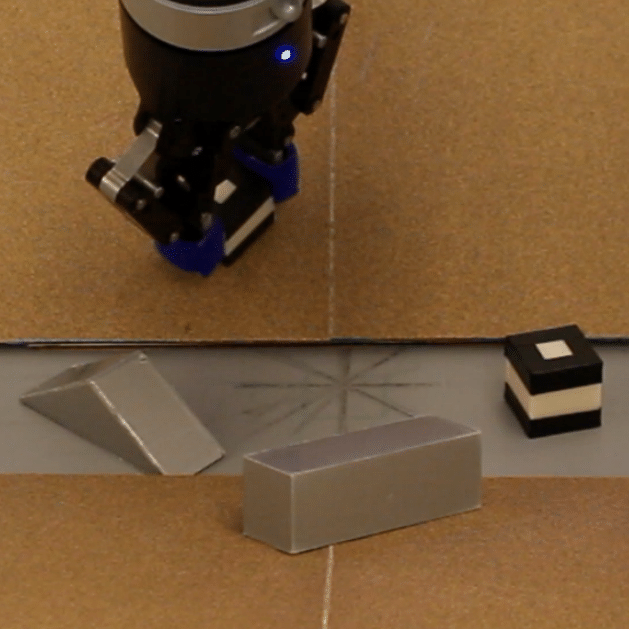}}
\subfloat[]{
\includegraphics[width=\robot]{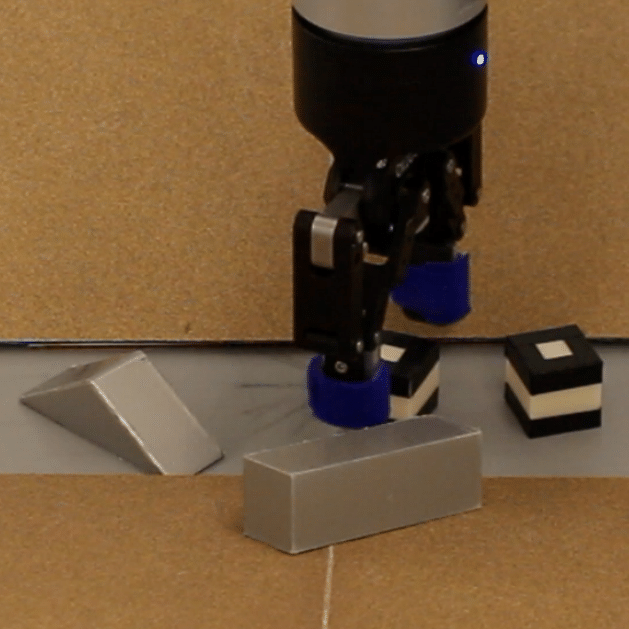}}
\subfloat[]{
\includegraphics[width=\robot]{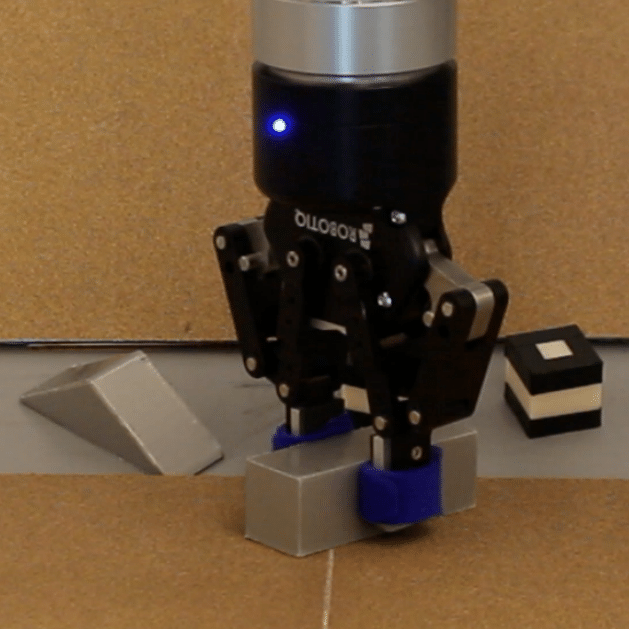}}
\subfloat[]{
\includegraphics[width=\robot]{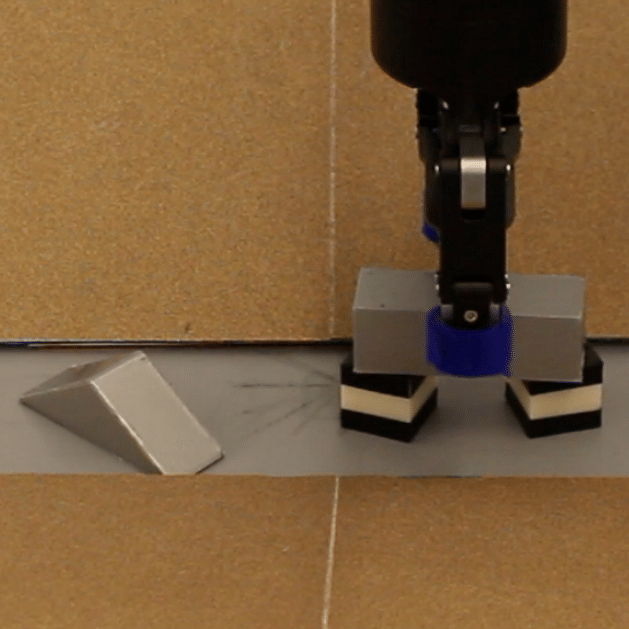}}
\subfloat[]{
\includegraphics[width=\robot]{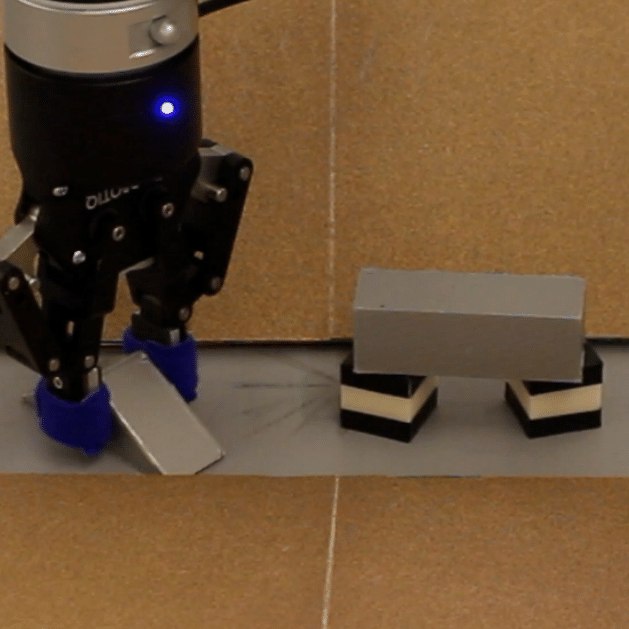}}
\subfloat[]{
\includegraphics[width=\robot]{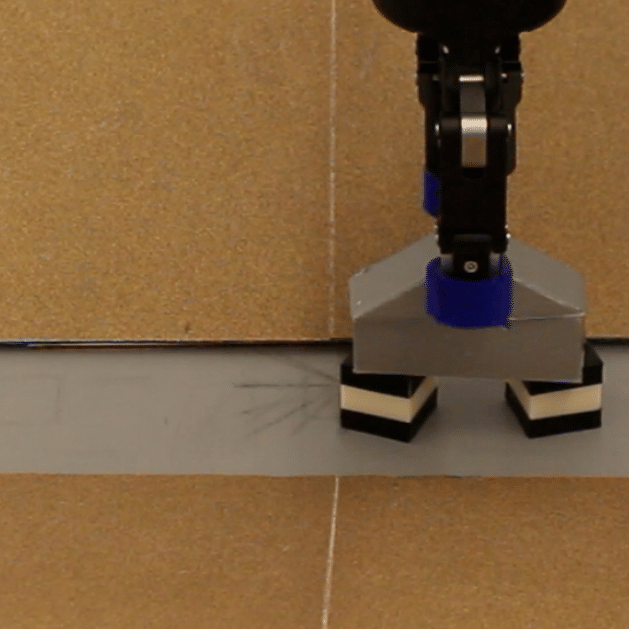}}
\caption{Example constructing the H3 structure.}
\label{fig:real_house3_demo}
\end{figure}

\begin{wrapfigure}[10]{r}{0.3\textwidth}
\centering
\vspace{-0.3cm}
\includegraphics[width=0.3\textwidth]{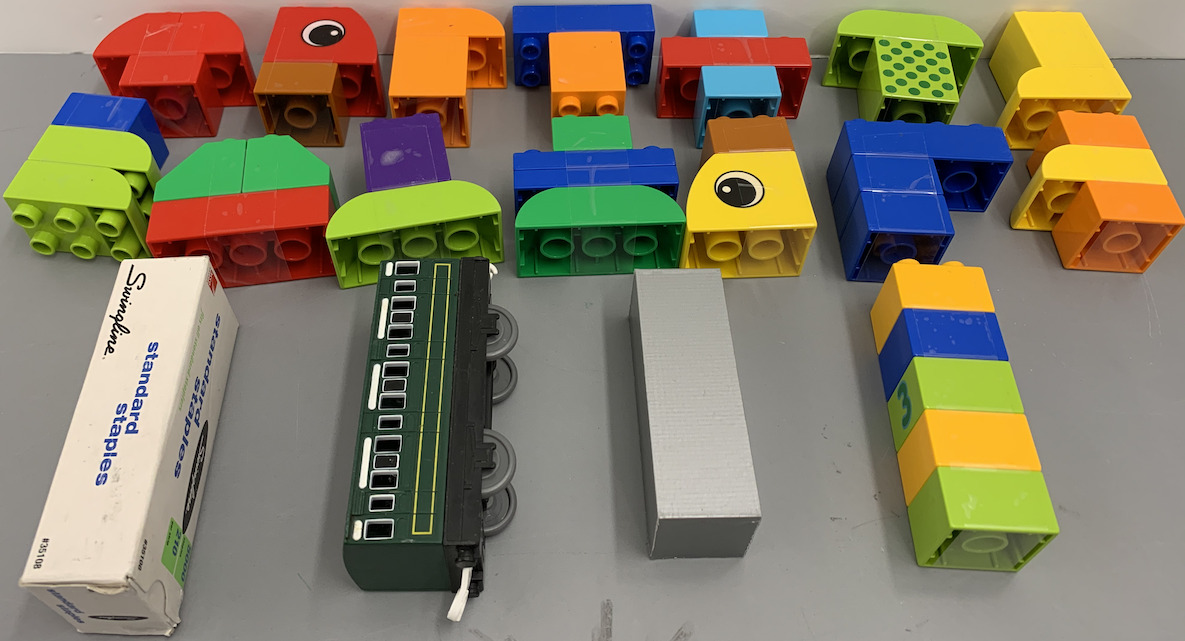}
\vspace{-0.6cm}
\caption{The unseen block shapes used to create structures in ImH2 and ImH3.}
\label{fig:realworl_im_obj}
\end{wrapfigure}

We tested the ASRSE3 SDQfD models for ImH2, H3, 4H1, ImH3, and H4 trained in PyBullet in Section~\ref{sect:exp_6d} on a Universal Robots UR5 arm with a Robotiq 2F-85 Gripper. An Occipital Structure sensor captures the scene image from a point looking down on the workspace. The slopes of the two ramps are 17deg and 28deg. For 4H1, H3, and H4, we use the objects with the same sizes in the simulator, while in ImH2 and ImH3 we use unseen shapes shown in Figure~\ref{fig:realworl_im_obj}. All other task parameters mirror simulation. Figure~\ref{fig:real_house3_demo} shows a run of task H3. For each block structure, we test the model for 40 episodes in 8 different orientations of the ramps. For each trial, if the robot successfully builds the target structure, it is recorded as a success. If the robot can't advance to the next intermediate structure in three pick-and-place action pairs, it is recorded as a failure. 

We demonstrate the success rate of each task, as well as the success rate for reaching the intermediate structures. Table~\ref{tab:robot_exp} shows the results of this experiment. In task 4H1, 36 of the 40 episodes succeed, demonstrating an 90\% success rate. In two of the four failures, the robot makes wrong decisions about place $xy$ positions. In the other two failures, the structure falls apart during unstable placing. In task H3, 39 of the 40 episodes succeed. In the only failure, the agent fails to predict the $xy$ position of the cube. In H4, 32 of the 40 episodes succeed, showing an 80\% success rate. In five of the eight failures, the cube fall off the brick during placing. The robot makes wrong decisions about pick $xy$ positions in the other three failures. In ImH2, 97.5\% of the episodes succeed. The failure is caused by a bad grasp point where it is too wide for the gripper. In the end, the robot succeeds 77.5\% in ImH3. In five failures the agent makes wrong decisions about the $xy$ position. In three failures the agent makes wrong decisions about the pick orientation. In one episode the structure is unstably built on the ramp.

\begin{table}
\newlength{\robottab}
\setlength{\robottab}{0.4in}
\centering
\subfloat
{
\begin{tabular}{|@{\hskip1pt}c@{\hskip1pt}|@{\hskip1pt}c@{\hskip1pt}|@{\hskip1pt}c@{\hskip1pt}|@{\hskip1pt}c@{\hskip1pt}|}
\hline
\shortstack{4H1 \\ \\ Block Structure} &
\includegraphics[width=\robottab]{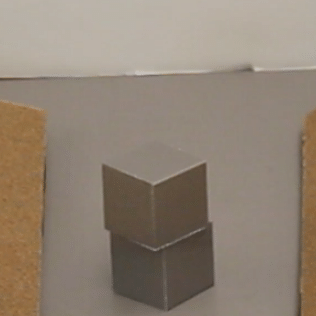} & \includegraphics[width=\robottab]{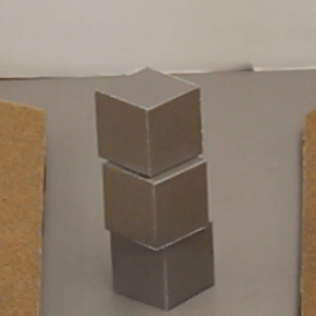} &
\includegraphics[width=\robottab]{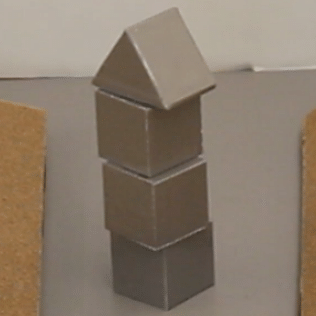} \\
\hline
Steps Needed & 2 & 4 & 6 \\
\hline
Success Rate & 100\% & 97.5\% & 90\% \\
& (40/40) & (39/40) & (36/40) \\
\hline
\hline
\shortstack{ImH2 \\ \\ Block Structure} &
\includegraphics[width=\robottab]{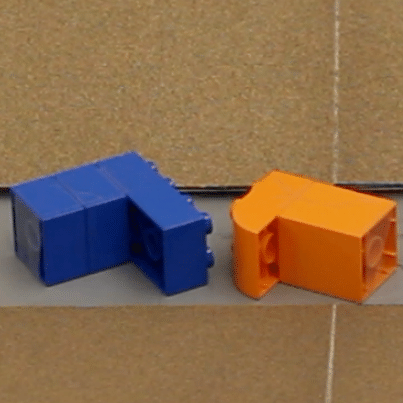} & \includegraphics[width=\robottab]{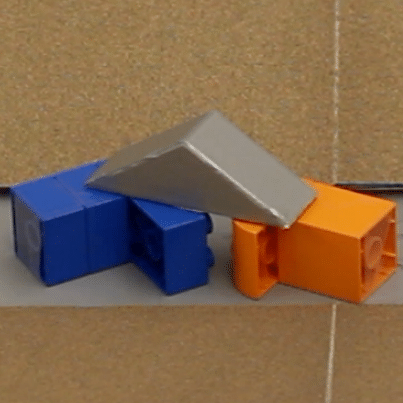} & \\
\hline
Steps Needed & 2 & 4 & - \\
\hline
Success Rate & 97.5\% & 97.5\% & - \\
& (39/40) & (39/40) & - \\
\hline
\end{tabular}
}
\subfloat
{
\begin{tabular}{|@{\hskip1pt}c@{\hskip1pt}|@{\hskip1pt}c@{\hskip1pt}|@{\hskip1pt}c@{\hskip1pt}|@{\hskip1pt}c@{\hskip1pt}|}
\hline
\shortstack{H3 \\ \\ Block Structure} &
\includegraphics[width=\robottab]{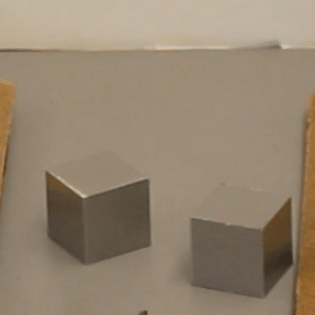} & \includegraphics[width=\robottab]{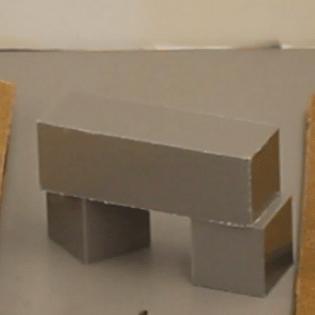} &
\includegraphics[width=\robottab]{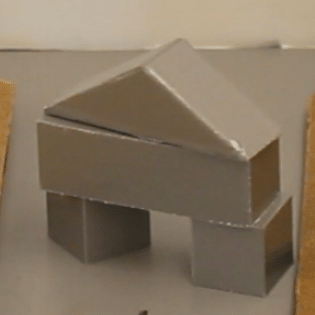} \\
\hline
Steps Needed & 2 & 4 & 6 \\
\hline
Success Rate & 97.5\% & 97.5\% & 97.5\% \\
& (39/40) & (39/40) & (39/40) \\
\hline
\hline
\shortstack{ImH3 \\ \\ Block Structure} &
\includegraphics[width=\robottab]{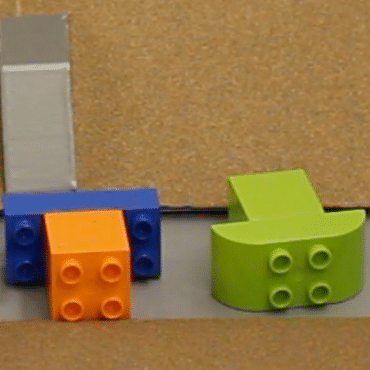} & \includegraphics[width=\robottab]{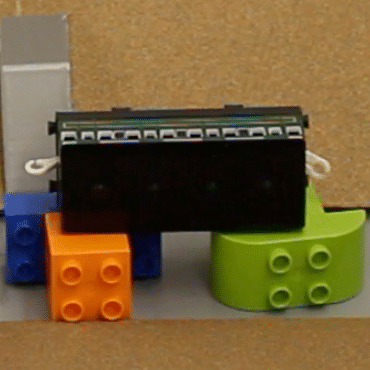} &\includegraphics[width=\robottab]{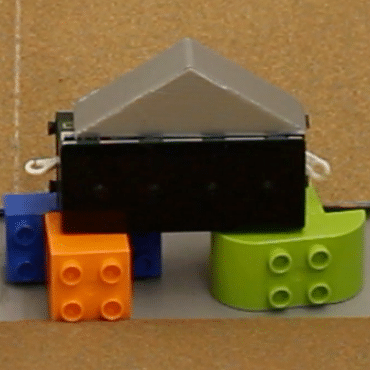} \\
\hline
Steps Needed & 2 & 4 & 6 \\
\hline
Success Rate & 85\% & 80\% & 77.5\% \\
& (34/40) & (32/40) & (31/40) \\
\hline
\end{tabular}
}\\
\vspace{-0.3cm}
\subfloat
{
\begin{tabular}{|@{\hskip1pt}c@{\hskip1pt}|@{\hskip1pt}c@{\hskip1pt}|@{\hskip1pt}c@{\hskip1pt}|@{\hskip1pt}c@{\hskip1pt}|@{\hskip1pt}c@{\hskip1pt}|@{\hskip1pt}c@{\hskip1pt}|}
\hline
\shortstack{H4 \\ \\ Block Structure} &
\includegraphics[width=\robottab]{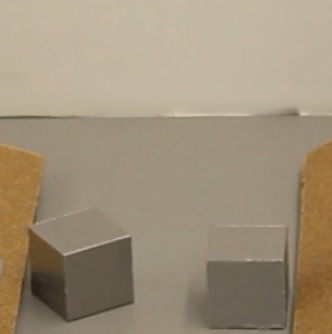} & \includegraphics[width=\robottab]{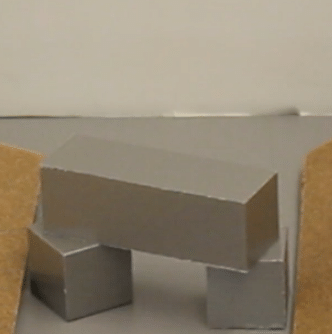} &
\includegraphics[width=\robottab]{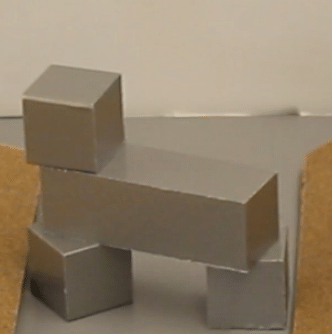} & \includegraphics[width=\robottab]{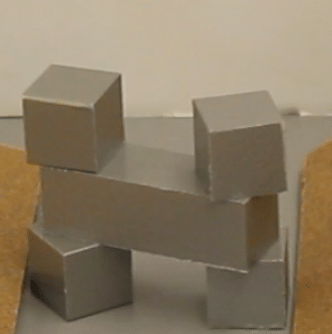} & \includegraphics[width=\robottab]{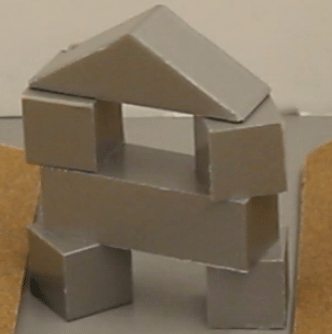}\\
\hline
Steps Needed & 2 & 4 & 6 & 8 & 10\\
\hline
Success Rate & 100\% & 100\% & 97.5\% & 82.5\% & 80\%\\
& (40/40) & (40/40) & (39/40) & (33/40) & (32/40)\\
\hline
\end{tabular}
}
\caption{Results of Robot Experiments}
\label{tab:robot_exp}
\vspace{-0.3cm}

\end{table}

\subsection{Conclusion and Limitations}

This paper makes two contributions. First, we propose ASRSE3, an augmented state representation for $SE(3)$ action spaces that enables us to solve manipulation problems with 6 DOF spatial action spaces. Second, we propose a variation of DQfD, SDQfD, that works more efficiently with the large action space. Our combined algorithm, ASRSE3 SDQfD, outperforms baseline methods in various experimental evaluations. It is able to learn policies that can assemble relatively complex block structures involving novel block shapes both in simulation and on a real robot. 

\noindent
\textbf{Limitations:} We expect that the approach extends beyond prehensile manipulation, as this has been done in related work~\cite{pushinggrasping}. A more fundamental limitation of this work is that it will need to be modified to accommodate the full 360 degrees of out-of-plane orientations -- the paper currently handles only $\pm30$ deg of orientation in $\phi$ and $\psi$. We might be able to accommodate a greater range of orientation by reasoning with respect to multiple viewpoints on the scene, but this is not clear. Another limitation is the need to define an expert planner for each new application domain. We expect that regression planning will extend well to many manipulation applications, but this has not been demonstrated. Finally, a critical area for future work is the extension from block structure building to more realistic domestic manipulation tasks. The fact that our approach can handle novel objects well suggests that this should be feasible, but it has yet to be demonstrated.




\clearpage
\acknowledgments{This work has been supported in part by the National Science Foundation (1724257, 1724191, 1763878, 1750649) and NASA (80NSSC19K1474).}


\bibliography{main}  

\clearpage

\appendix
\section{State and Action Encoding}
\label{appen:state_action_enc}
The state space contains the top-down scene image $I\in\mathbb{R}^{90\times90}$, the ``in-hand'' image $H\in\mathbb{R}^{3\times24\times24}$ and a bit $g$ describing the previous gripper action (i.e. if the gripper is holding an object). We apply a Perlin noise~\cite{perlin_noise} on the scene image to increase the generalizability of the trained model. 

In the FCN baseline methods, $I$ needs to be rotated for encoding $\theta$. To prevent information loss caused by the rotation, $I$ is padded with 0 to the size of $128\times128$ before rotation and passing to the network. The padding is removed later in the output. To ensure a fair comparison, we do the same padding in ASRSE3 though it is not necessary for our approach.

\begin{figure}[h]
\centering
\subfloat[]{
\includegraphics[width=0.2\linewidth]{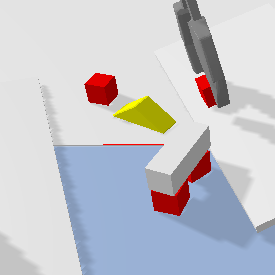}
}
\subfloat[]{
\includegraphics[width=0.2\linewidth]{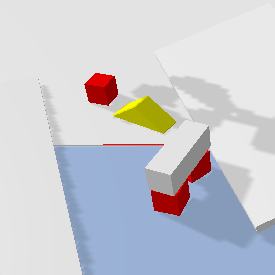}
}
\subfloat[]{
\includegraphics[width=0.2\linewidth]{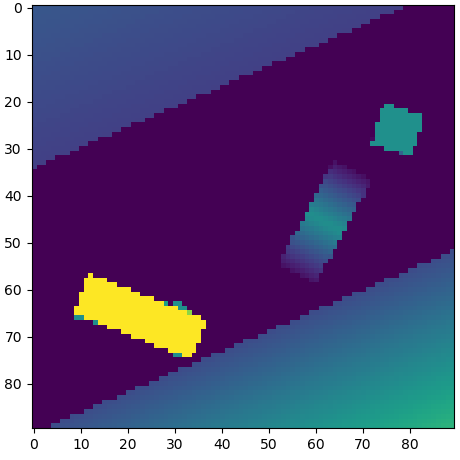}
}\\
\subfloat[]{
\includegraphics[width=0.15\linewidth]{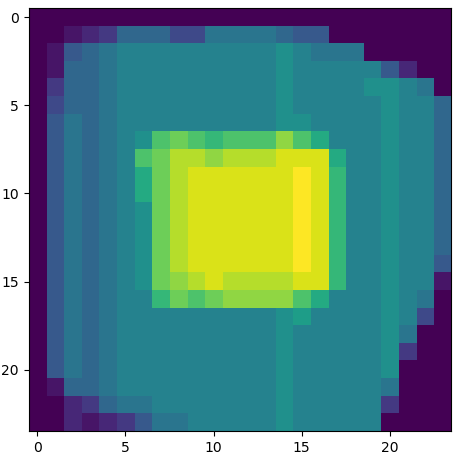}
}
\subfloat[]{
\includegraphics[width=0.15\linewidth]{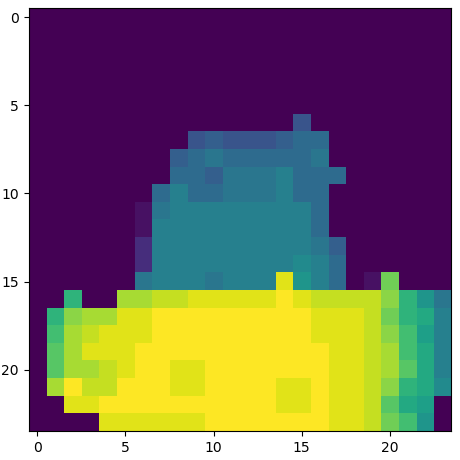}
}
\subfloat[]{
\includegraphics[width=0.15\linewidth]{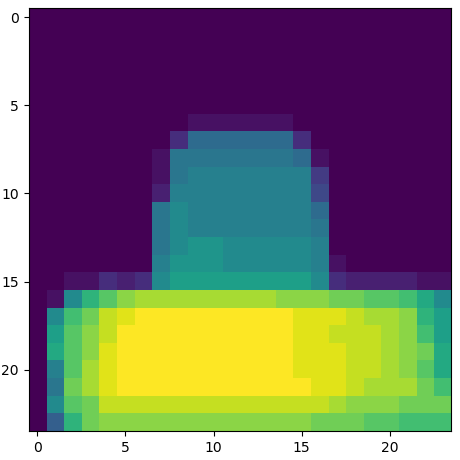}
}
\caption{An example of the state space. (a) The pick action at the previous time step. (b) The workspace scene. (c) The top-down observation $I$. (d)-(f) The in-hand image $H$. There is also a bit $g$ indicating if the gripper is holding an object. In this example it is 1 because the gripper is holding a cube block.}
\label{fig:state_example}
\end{figure}

The in-hand image is an orthographic projection of a partial point cloud at where the last pick occurred. If the last action was a $\textsc{place}$, then $H_t$ is set to zero. If the last action was a $\textsc{pick}$, then we first generate a point cloud based on the previous scene image $I_{t-1}$, and transform it into the end-effector frame of the previous pick action target pose. The point cloud is then voxelized into an occupancy grid $V\in\mathbbm{1}^{24\times24\times24}$, where each cell in $V$ is a binary value representing if that cell is occupied in the point cloud. Finally, $H_t\in\mathbb{Z}^{3\times24\times24}$ is set to an orthographic projection of $V$ by taking the sum along the three dimensions in $V$. The purpose of the in-hand image is to retain information about the shape and pose of the object currently grasped by the robot. Figure~\ref{fig:state_example} shows an example of the state space. When ASRSE3 is applied in $x, y, \theta$ action space, $H$ is simplified into an image crop of $I$ centered at $x, y$ rotated by $\theta$. 

The encoding function $f_4(s, a_1, a_2, a_3)$ and $f_5(s, a_1, a_2, a_3, a_4)$ use a similar orthographic projection generation procedure as the in-hand image $H$, while instead of using the previous scene image and transforming the point cloud into the frame of the previous pick action pose, $f_4$ and $f_5$ use the current scene image and the point cloud is transformed into the frame of the partially selected action $(a_1, a_2, a_3)$ or $(a_1, a_2, a_3, a_4)$ where $a_1=a_{xy}, a_2=a_\theta, a_3=a_z, a_4=a_\phi$.

The partial action space $A_{xy}$ has a same size as the pixel size of $I$, i.e. each pixel in $I$ is a potential action in $A_{xy}$. The workspace has a size of $0.3\rm{m} \times 0.3\rm{m}$, therefore, each pixel corresponds to a $0.003\rm{m}\times0.003\rm{m}$ region in the scene. This is an upper bound on the position accuracy of $a_{xy}$. $A_\theta$ is discretized into 8 values from 0 to $\frac{7}{8}\pi$. ($[\pi, 2\pi]$ is omitted because the gripper is symmetrical about the $z$ axis). $A_\phi$ and $A_\psi$ are discretized into 7 values from $-\frac{\pi}{6}$ to $\frac{\pi}{6}$. $A_{z}$ is discretized into 16 values from 0.02m to 0.12m. The DDPG baseline uses continues action spaces within the same range. 

\section{Network Model Architectures}
\label{appen:network}
\subsection{Model architecture used by ASRSE3}

\begin{figure}
    \centering
    \includegraphics[width=\linewidth]{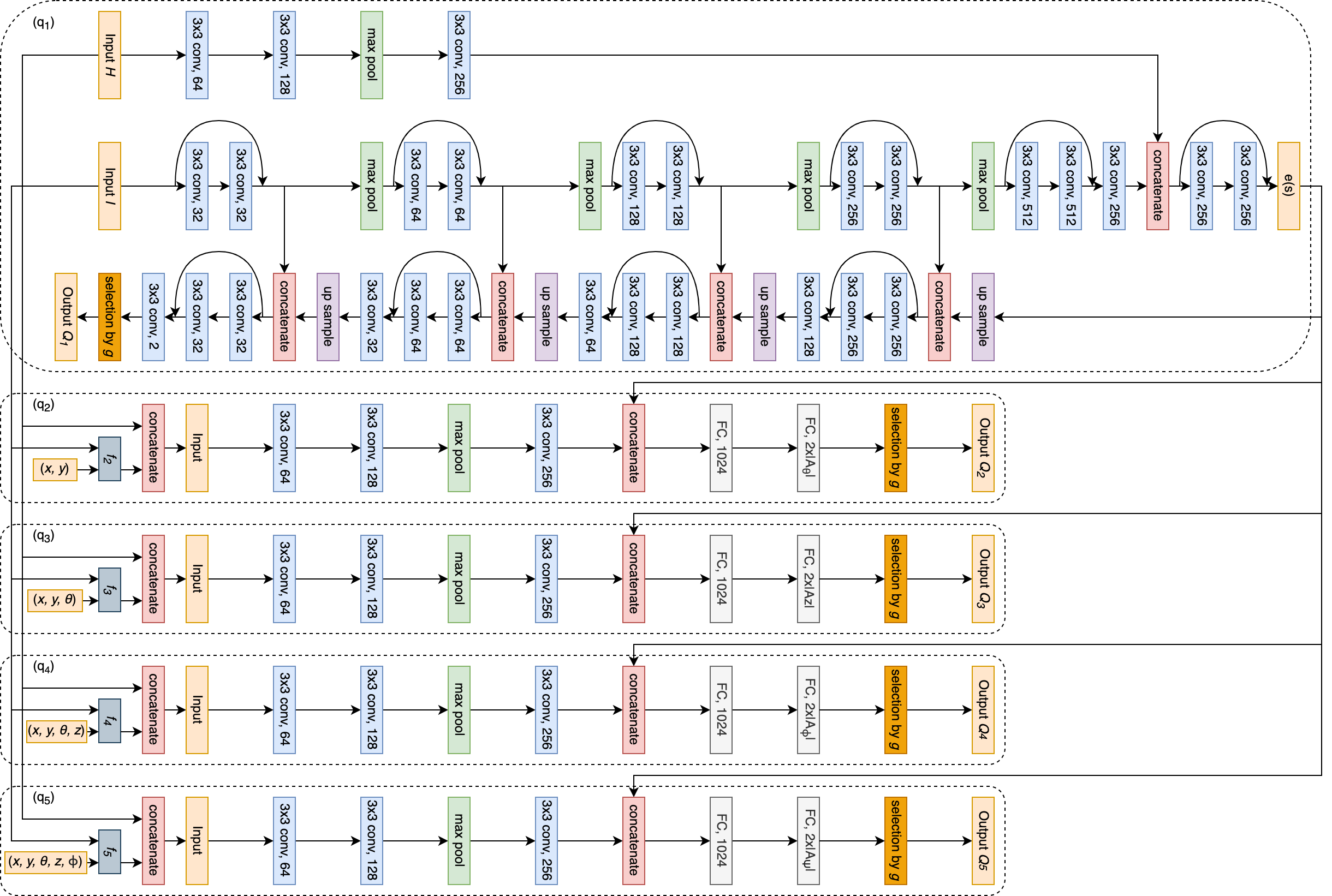}
    \caption{Our network architecture for ASRSE3. The curved connection represents the shortcut connection for residual learning~\cite{resnet}. ReLU nonlinearity is omitted in the diagram}
    \label{fig:network}
\end{figure}

In order to use Neural Networks for the augmented state representation, there are two main challenges. First, $q_1$ needs to be an FCN to implement pixel level labeling, and it needs to reason about both local information and global information to generate $Q$ values. Second, all $Q$ functions share the common input of state $s$, training separate encoders in each $Q$ net will be redundant. To address those problems, we use an architecture showed in Figure~\ref{fig:network}. UNet~\cite{unet} is a good fit for $q_1$ to solve our first problem because it first expands the whole image into small resolutions in the encoding phase so that the convolutional filters can capture global information. In the decoding phase, the concatenation between the feature map in the main path and the feature map from the previous level guarantees that the local information is not lost. To incorporate the in-hand image $H$ in our approach, we train a separate encoder that encodes $H$ into a feature map and concatenates it with the bottle neck feature map of the UNet. Figure~\ref{fig:network} (top) shows the detailed UNet architecture. For $q_2$, $q_2$, $q_4$, and $q_5$, we train an encoder that encodes the image from $f_2 - f_5$ into feature maps. This feature map is then concatenated with the bottle neck feature map $e(s)$ from the UNet encoder. This concatenation enables the network to reason about state information without re-training the state encoder. Figure~\ref{fig:network} (bottom) shows the detailed architecture of $q_2$ to $q_5$. In all five networks, the $Q$ values for $\textsc{pick}$ and $\textsc{place}$ are encoded by separate heads. The gripper status $g$ selects which channel to output.

\subsection{Model architecture used in FCN baselines}
For FCN baseline methods, we use the same architecture as the stated $q_1$ network. Same as the prior work~\cite{zeng2018roboticpick-and-place, pushinggrasping}, in such approach, $|A_\theta|$ copies of $I$ is created, corresponding to each $a_\theta\in A_\theta$. For each copy, the whole image is rotated by $a_\theta$ to represent $\theta$. After forward passing each of the $|A_\theta|$ copies, the output $Q$ maps are rotated back respectively by the corresponding $a_\theta$. 

\subsection{Model architecture used in DDPG}
In the actor model, we use a ResNet-34 backbone for the input $I$ and a standard CNN for input $H$. The outputs of those two are flattened and sent to the fully connected layers with Tanh activation. The output is two 3-vectors, representing the pick and place action of each of the $x, y, \theta$ dimensions. Similar to our other models, $g$ is used to select which vector to use. The action value is scaled to the range of each action dimensions at execution. The critic model also uses a ResNet-34 backbone for encoding $I$ and a standard CNN for encoding $H$, but it also has an FC layer for encoding the action. The flattened feature maps of $I$ and $H$ are concatenated with the feature vector of the action, then fully connected layers with ReLU activation are used for generating the $Q$ value. Similarly, the critic network outputs two values for pick and place respectively, and $g$ is used to select the actual value.

\section{$n$-step return in augmented MDP}
\label{appen:n-step}
\begin{figure}[t]
\centering
\includegraphics[width=\linewidth]{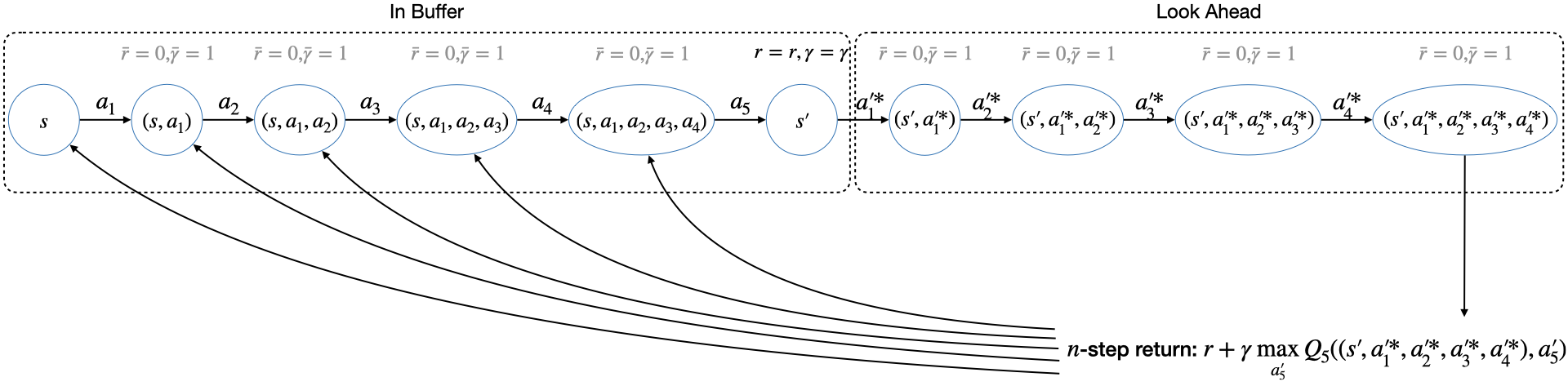}
\caption{The $n$-step return in the augmented state representation}
\label{fig:n-step}
\end{figure}
Recall that we need to learn five $Q$ functions, $Q_1, Q_2, Q_3, Q_4, Q_5$, by minimizing a loss $\mathcal{L}_i = \mathbb{E} \big(y_i - Q_i(\bar{s},\bar{a}) \big)^2$ for a TD target $y_i$. To do this, we need to calculate a TD estimate $y_i$ of return for the augmented MDP $\bar{\mathcal{M}}$. Given a transition $(s, a_1, a_2, a_3, a_4, a_5, s', r)$, the $1$-step TD return is:
\begin{equation}
\begin{split}
y_1 &= \bar{r}+\bar{\gamma}\max_{a^0_2} Q_2((s, a_1), a^0_2)\\
&=\max_{a^0_2} Q_2((s, a_1), a^0_2)
\end{split}
\end{equation}
\begin{equation}
\begin{split}
y_2 &= \bar{r}+\bar{\gamma}\max_{a^0_3} Q_3((s, a_1, a_2), a^0_3)\\
&=\max_{a^0_3} Q_3((s, a_1, a_2), a^0_3)
\end{split}
\end{equation}
\begin{equation}
\begin{split}
y_3 &= \bar{r}+\bar{\gamma}\max_{a^0_4} Q_4((s, a_1, a_2, a_3), a^0_4)\\
&=\max_{a^0_4} Q_4((s, a_1, a_2, a_3), a^0_4)
\end{split}
\end{equation}
\begin{equation}
\begin{split}
y_4 &= \bar{r}+\bar{\gamma}\max_{a^0_5} Q_4((s, a_1, a_2, a_3, a_4), a^0_5)\\
&=\max_{a^0_5} Q_5((s, a_1, a_2, a_3, a_4), a^0_5)
\end{split}
\end{equation}
\begin{equation}
\label{eqn:1-step_q5}
y_5=r+\gamma\max_{a'_1}Q_1(s', a'_1)
\end{equation}

However, we have found that our ASRSE3 learns faster with an $n$-step return where each of the five value functions is trained with a TD return of 
\begin{equation}
y_1=y_2=y_3=y_4=y_5 = r+\gamma\max_{a'_5}Q_5((s', a'^*_1, a'^*_2, a'^*_3, a'^*_4), a'_5)
\end{equation}
where 
\begin{align}
a'^*_1&=\argmax_{a'_1}Q_1(s', a'_1)\\
a'^*_2&=\argmax_{a'_2}Q_2((s, a'^*_1), a'_2)\\
a'^*_3&=\argmax_{a'_3}Q_3((s, a'^*_1, a'^*_2), a'_3)\\
a'^*_4&=\argmax_{a'_4}Q_4((s, a'^*_1, a'^*_2, a'^*_3), a'_4)
\end{align}
$a'^*_1, a'^*_2, a'^*_3, a'^*_4$ are the greedy actions selected by $Q_1$, $Q_2$, $Q_3$, and $Q_4$. The reward $\bar{r}=0$ and discount factor $\bar{\gamma}=1$ for the augmented states are omitted in the equation. This process is illustrated in Figure~\ref{fig:n-step}. Note that the states after $s'$ are not experienced by the agent -- it is created by augmenting $s'$ with the greedy actions. This $n$-step return could be viewed as a 9-step return for $Q_1$, an 8-step return for $Q_2$, a 7-step return for $Q_3$, a 6-step return for $Q_4$, and a 5-step return for $Q_5$. 


The $n$-step return propagates the outcome of the future states faster. Still, it has two potential problems: the high variance due to the stochasticity in the transitions, and the incorrect optimal return value caused by the sub-optimal action selections. In our approach, however, the deterministic transition dynamics between the intermediate states and the greedy policy during training conquer the first problem. The second problem is moderated by the imitation learning aspect of our work because the actions in the expert transitions could be viewed as optimal.

\begin{figure}[t]
\centering
\subfloat[5H1]
{
\includegraphics[width=0.2\linewidth]{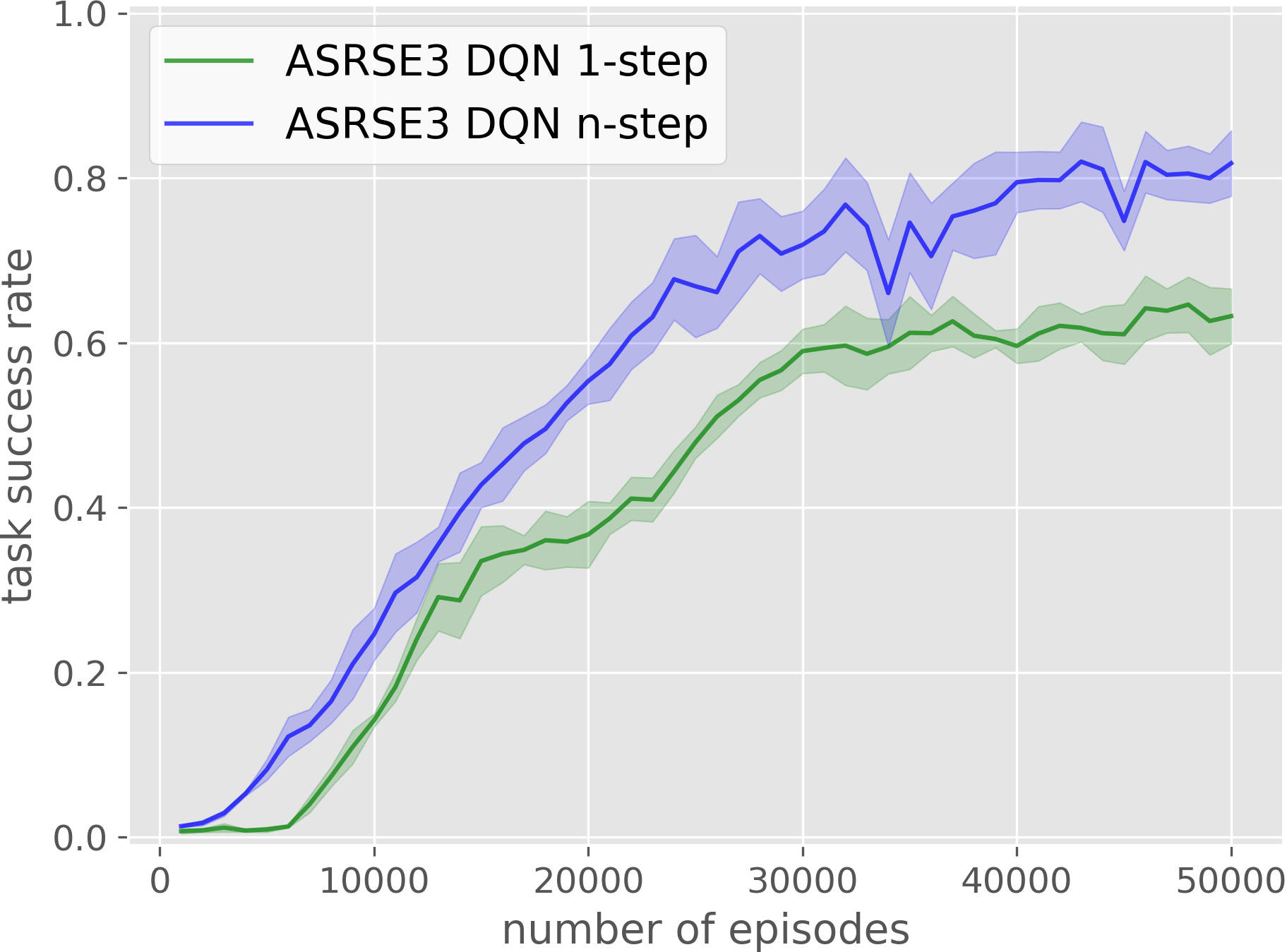}\hfill
}
\subfloat[H4]
{
\includegraphics[width=0.2\linewidth]{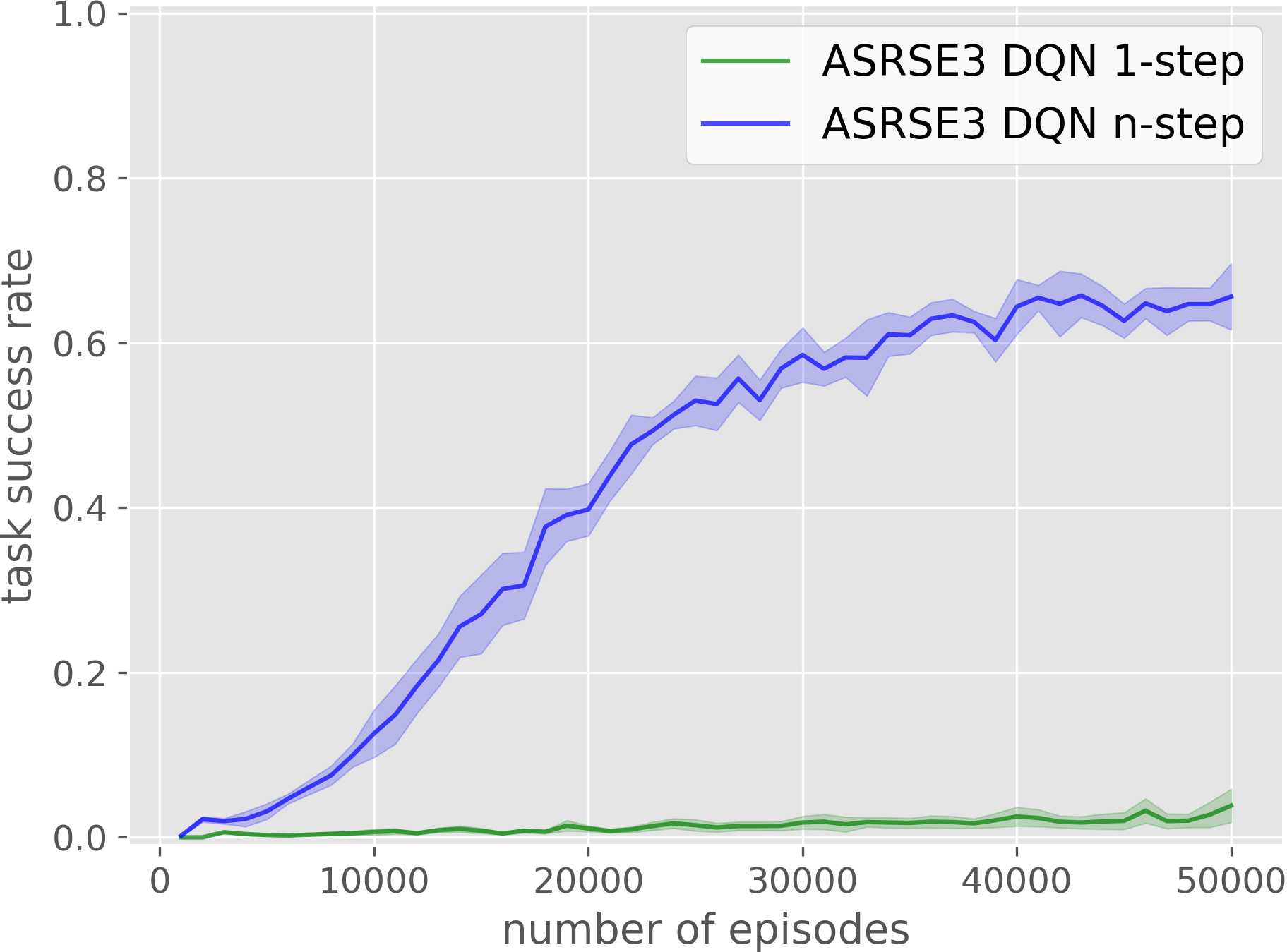}\hfill
}
\subfloat[ImDis]
{
\includegraphics[width=0.2\linewidth]{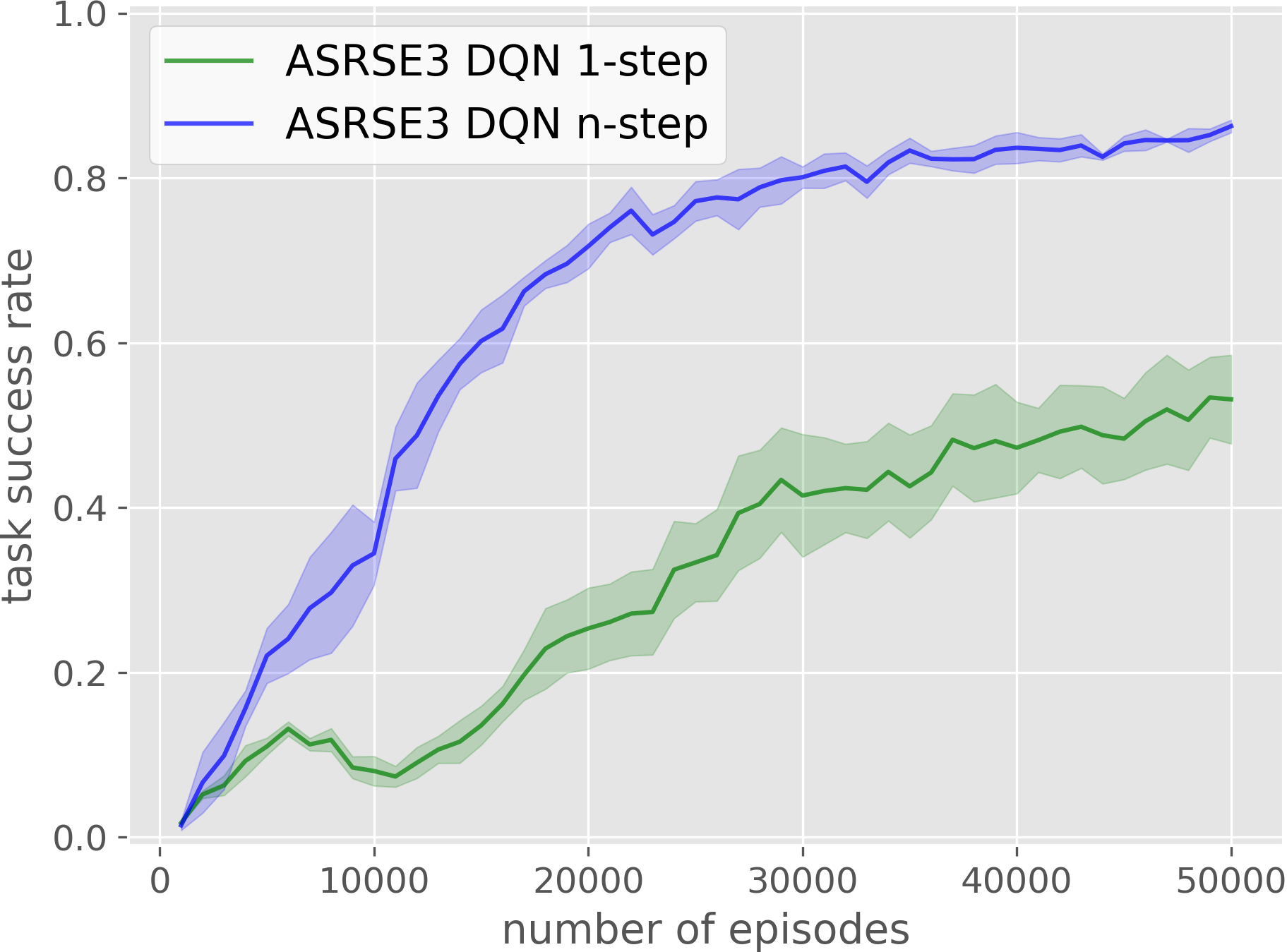}\hfill
}
\subfloat[ImRan]
{
\includegraphics[width=0.2\linewidth]{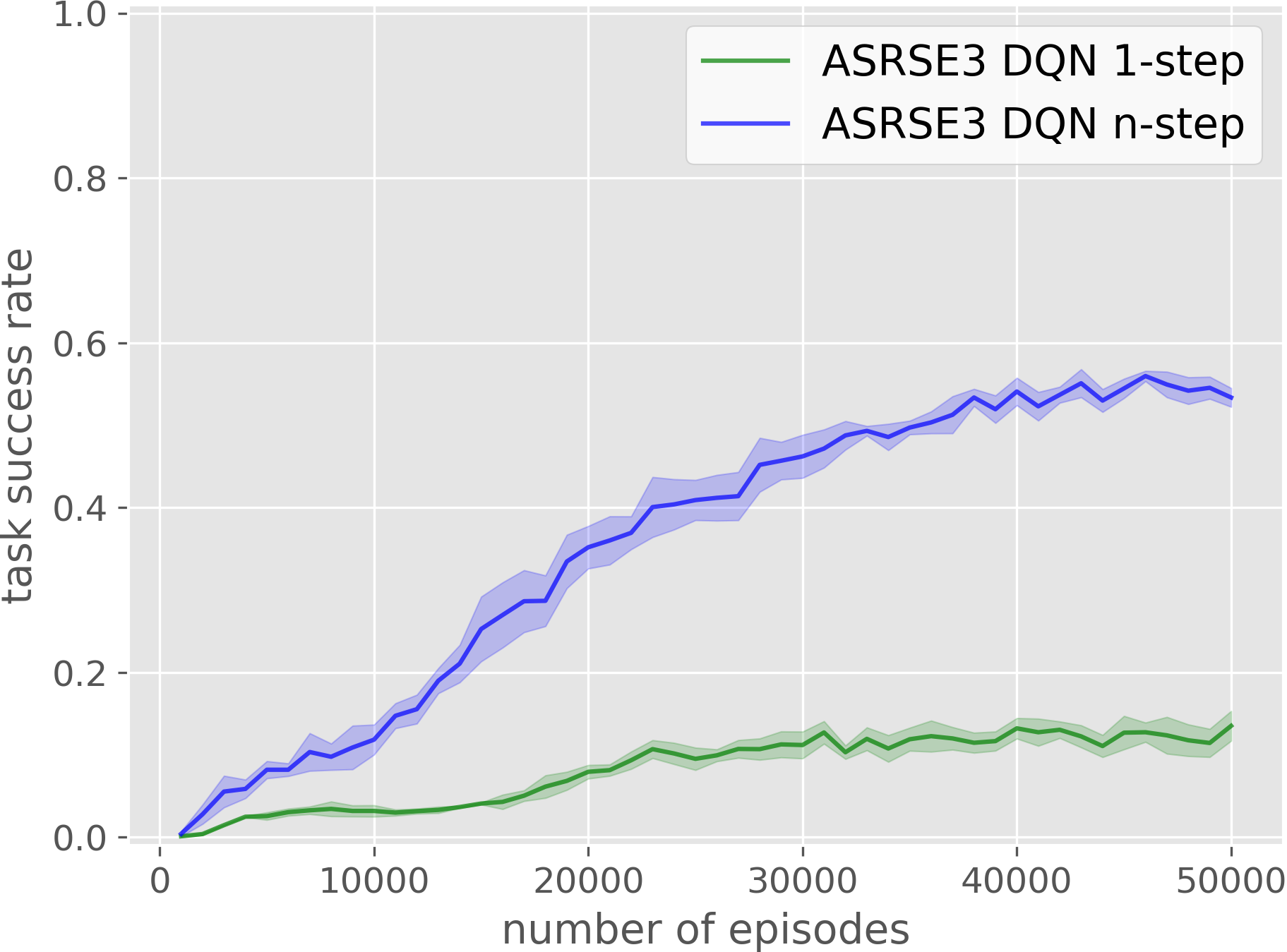}\hfill
}
\caption{Learning curves of ASRSE3 DQN with $1$-step return and $n$-step return in $(x, y, \theta)$ action space. All algorithms have a pretrain phase of 10k training steps over deconstruction expert transitions. The results are averaged over 4 runs and plotted with a window of 1000 episodes.}
\label{fig:exp_1step}
\end{figure}

\begin{figure}[t]
\centering
\subfloat[4S]
{
\includegraphics[width=0.2\linewidth]{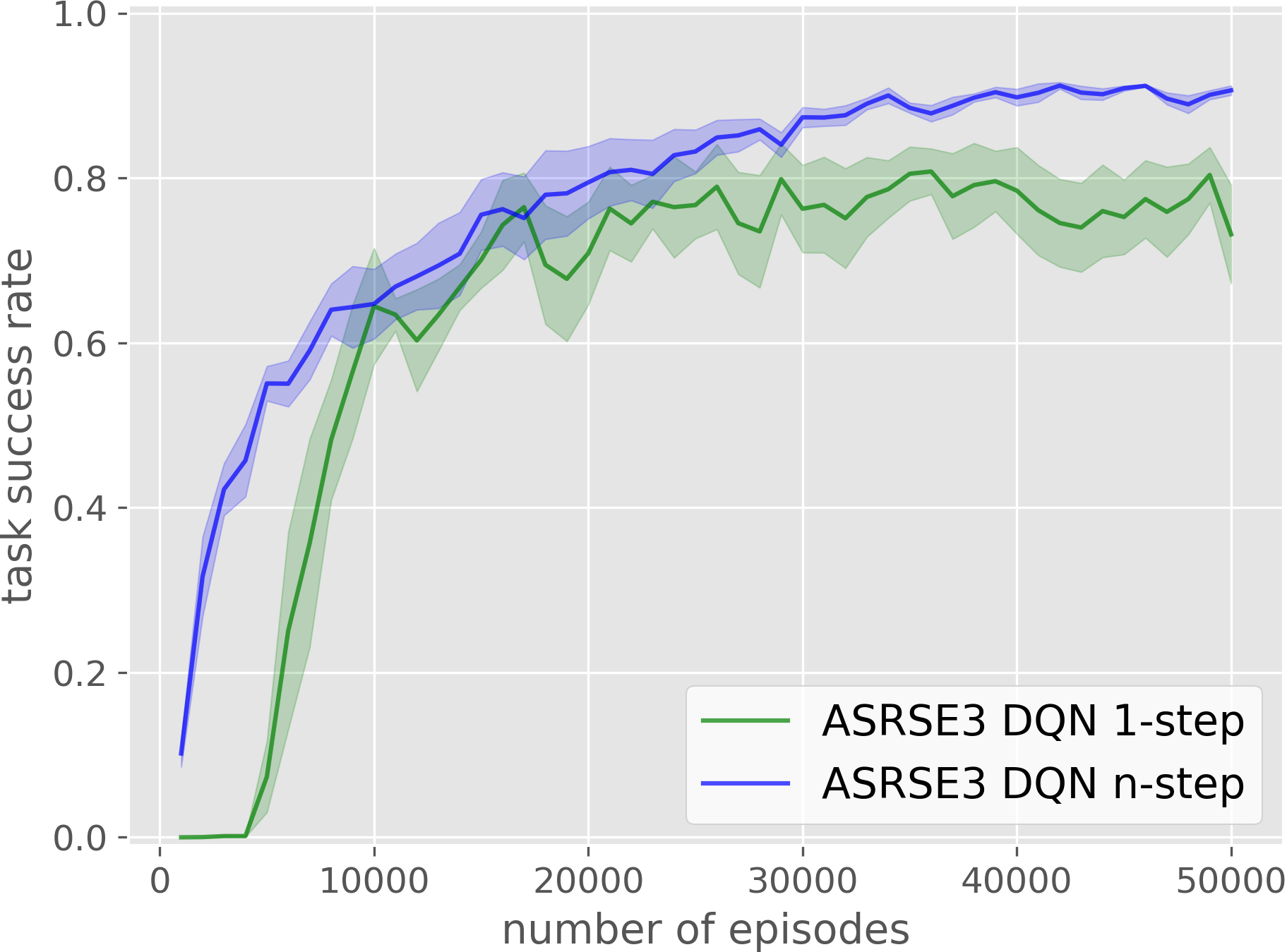}\hfill
}
\subfloat[4H1]
{
\includegraphics[width=0.2\linewidth]{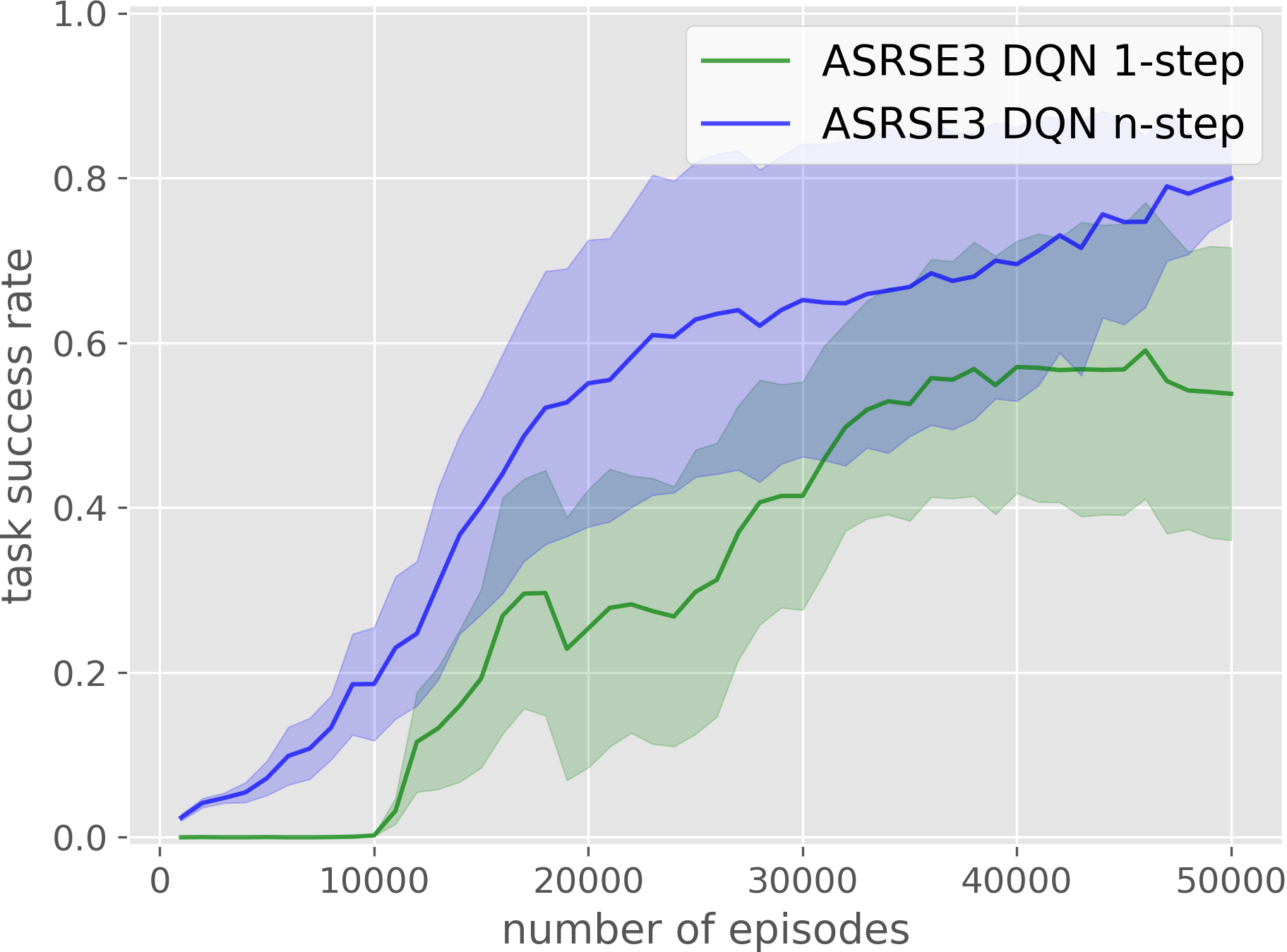}\hfill
}
\subfloat[H2]
{
\includegraphics[width=0.2\linewidth]{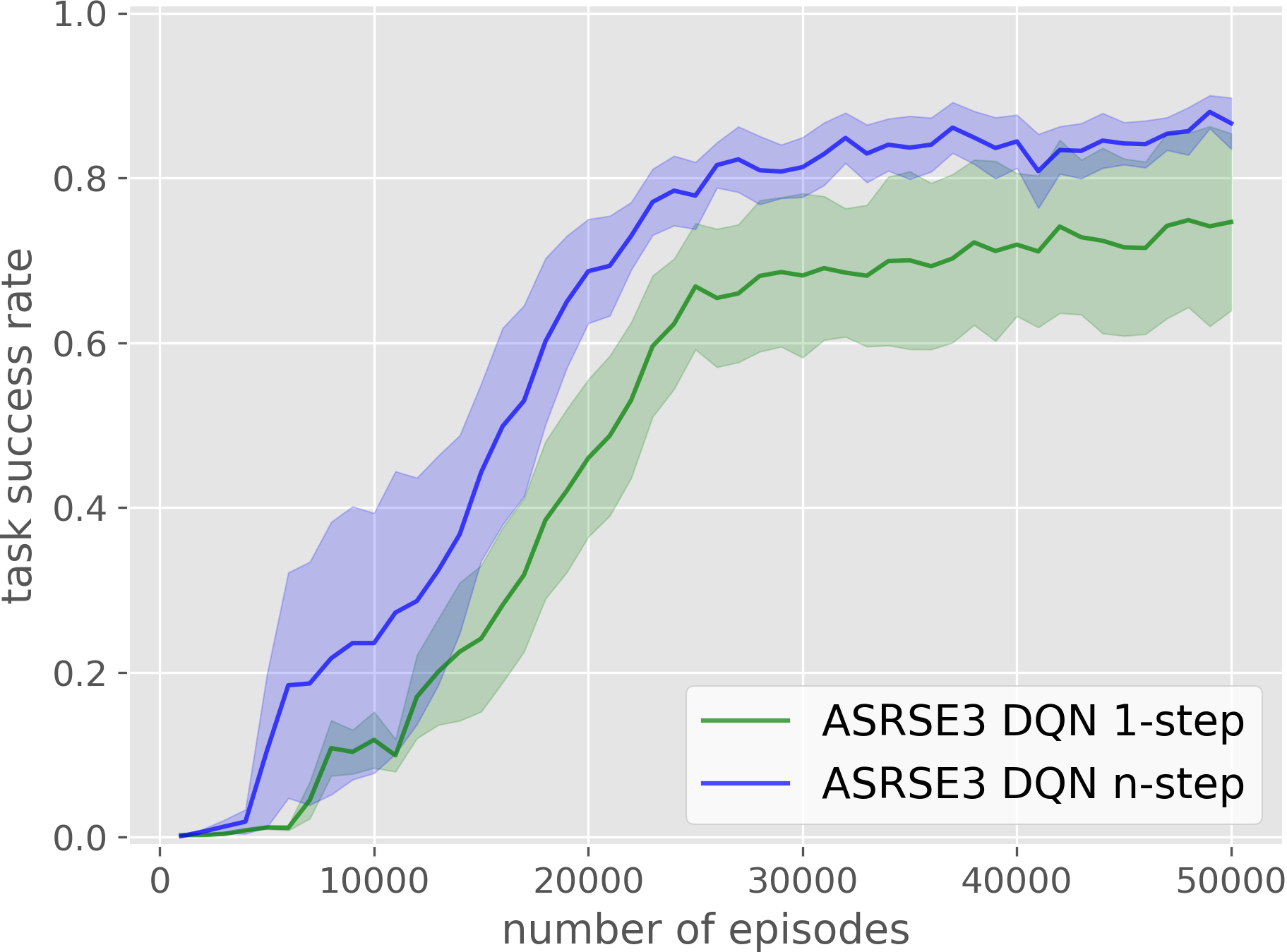}\hfill
}
\subfloat[ImH2]
{
\includegraphics[width=0.2\linewidth]{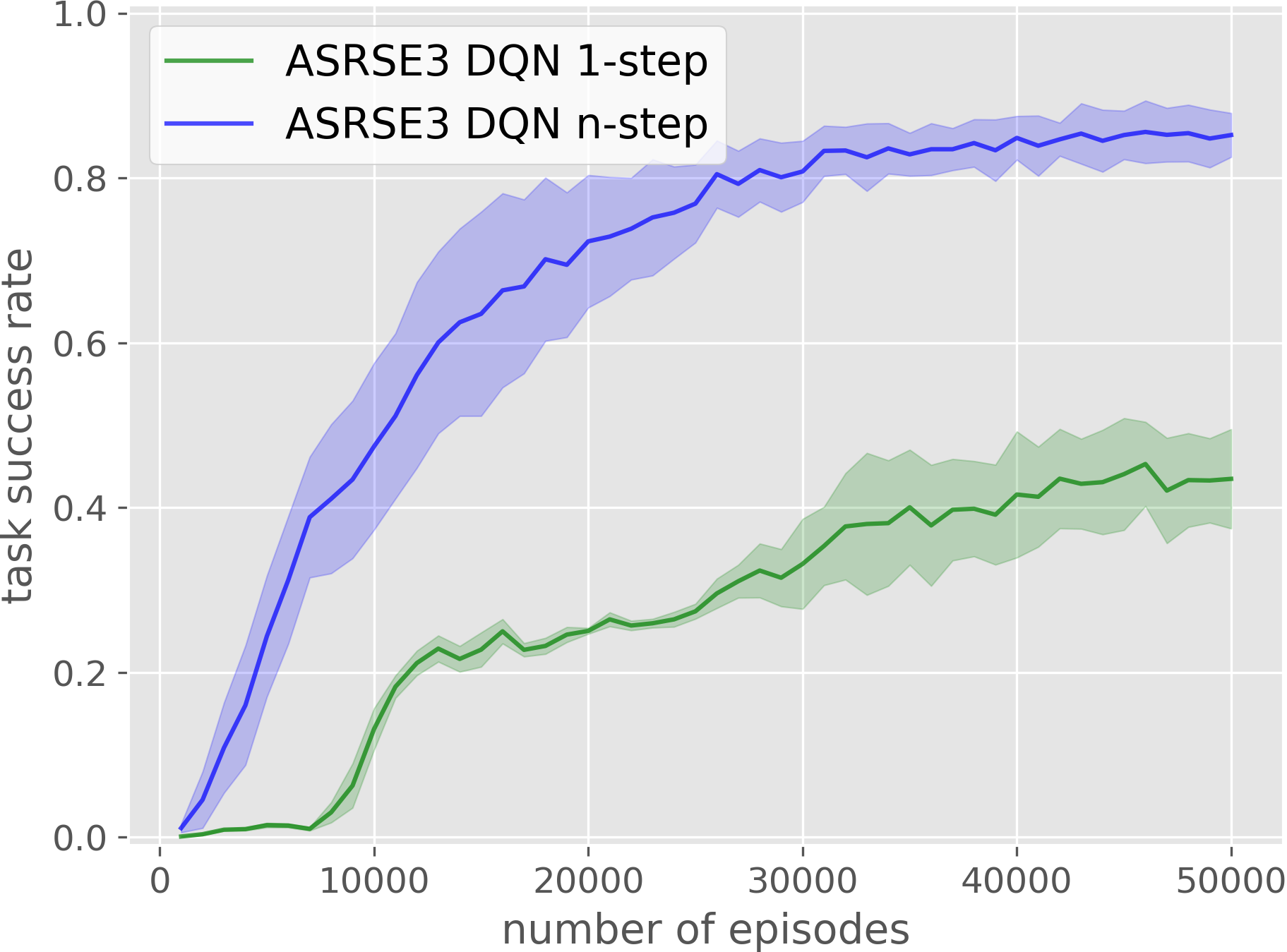}\hfill
}
\caption{Learning curves of ASRSE3 DQN with $1$-step return and $n$-step return in 6-DOF action space. All algorithms have a pretrain phase of 10k training steps over deconstruction expert transitions. The results are averaged over 4 runs and plotted with a window of 1000 episodes.}
\label{fig:exp_1step_6d}
\end{figure}

To evaluate the $n$-step return, we perform an experiment that compares the $1$-step return and $n$-step return versions of ASRSE3 DQN (with pretraining) in 5H1, H4, ImDis, and ImRan (see Figure~\ref{fig:envs}) tasks in the $(x, y, \theta)$ action space on a flat workspace (same setting of Section~\ref{sect:exp_sdqfd}). Both DQN variations have a pretraining phase of 10k training steps using expert transitions, then starts gathering self-play data (see Appendix~\ref{appen:algorithm_detail}).
Figure~\ref{fig:exp_1step} shows the result. The $n$-step return version always outperforms the $1$-step return version. Note that the $n$-step return is especially helpful in challenging tasks with more required action steps (e.g. H4) because it can propagate the positive outcome to the early states faster. 1-step DQN can barely learn in H4 task while $n$-step DQN can reach more than 60\% success rate.

We then perform a similar experiment to compare the $1$-step return and $n$-step return versions of ASRSE3 DQN (with pretraining) in the 6-DOF action space. The experiment is performed in 4S, 4H1, H2, ImH2 (see Figure~\ref{fig:envs}) tasks in the ramp workspace (same setting of Section~\ref{sect:exp_6d}). Figure~\ref{fig:exp_1step_6d} shows the result. Similar to the previous experiment in the $(x, y, \theta)$ action space, $n$-step return outperforms $1$-step return in all environments in the 6-DOF action space.

\section{Generating expert trajectories via structure deconstruction}
\label{appen:decons}

\begin{figure}[t]
\centering
\includegraphics[width=\linewidth]{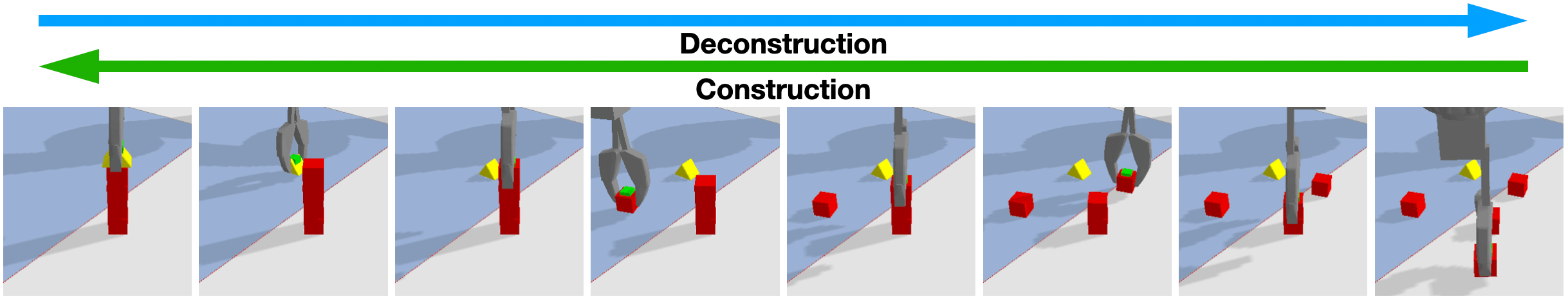}
\caption{An example of getting construction transitions from deconstruction. From left to right: an episode of deconstructing the block structure finished by the expert. From right to left: after reversing the episode, we get an expert construction episode.}
\label{fig:deconstruction}
\end{figure}

Since our method has access to the full state of the simulator during training, one intuitive way for generating expert demonstrations for IL algorithms is to hand code an expert policy that reads the block positions and builds a state machine for each phase during construction. However, this means that we need to hand code a new expert for each new block structure to build. \citet{form2fit} introduces a method that learns kit assembly from disassembly by reversing the recorded disassembly transitions. Disassembly is intuitively easier than assembly because no matching between objects and kits is required. Inspired by this work, we find that a similar methodology can be used in block construction environments. We code an expert deconstruction policy that simply picks up the highest block and places it on the ground with a random position and orientation. By reversing the deconstruction episode, we can acquire an expert construction episode. Moreover, this expert deconstruction policy can be applied to all block construction domains in our experiments. Fig~\ref{fig:deconstruction} shows an example of getting construction transition from deconstruction.

\section{Algorithm Details}
\label{appen:algorithm_detail}
In SDQfD, DQN with pretraining, ADET, and DQfD, there are two training phases: pretraining and self-play. An expert buffer of 50k deconstruction transitions is loaded, and the agent is trained solely on those expert data for 10k training steps in the pretraining phase. After the pretraining phase, the agent starts gathering self-play data stored in a separate self-play buffer. During training in the self-play phase, the agent sample 50\% of the data from the expert buffer, and 50\% from the self-play buffer. The second phase lasts for 50k episodes. In SDQfD, DQfD, and ADET, the margin loss (SDQfD, DQfD) and cross entropy loss (ADET) is only applied to the expert transitions. In BC, we train the policy network using the cross entropy loss, and the policy is a deterministic policy using the max function over the trained network. BC methods are trained for 50k training steps using the expert transitions and tested for 1000 episodes. Algorithm~\ref{alg:asrse3_sdqfd} shows the detailed algorithm of ASRSE3 SDQfD.

\begin{algorithm}[ht]
\begin{algorithmic}
\STATE{Initialize $Q$ networks $q_1, q_2, q_3, q_4, q_5$, target networks $\bar{q_1}, \bar{q_2}, \bar{q_3}, \bar{q_4}, \bar{q_5}$, expert buffer $\mathcal{D}_e$, self-play buffer $\mathcal{D}_s$, margin loss weight $w$, pretraining steps $M$, maximum episodes $N$}
\STATE{Load expert buffer $\mathcal{D}_e$}
\FOR{$i=1,M$}
\STATE{Sample minibatch $(s, a_1, a_2, a_3, a_4, a_5,  s', r)$ from $\mathcal{D}_e$}
\STATE{Calculate TD target $y_{[1-5]} = r + \gamma \max \bar{q_5}(e(s'), H', f_5(s',a'^*_1,a'^*_2, a'^*_3, a'^*_4))$ \\where:\\ $a'^*_1=\argmax \bar{q_1}(s')$ \\ $a'^*_2=\argmax \bar{q_2}(e(s'), H', f_2(s', a'^*_1))$ \\ $a'^*_3=\argmax \bar{q_3}(e(s'), H', f_3(s', a'^*_1, a'^*_2))$ \\ $a'^*_4=\argmax \bar{q_4}(e(s'), H', f_4(s',a'^*_1,a'^*_2, a'^*_3))$\\ $e(s')$ is from the forward pass of $\bar{q_1}(s')$}
\STATE{Calculate $\mathcal{L}_{TD1}, \mathcal{L}_{TD2}, \mathcal{L}_{TD3}, \mathcal{L}_{TD4}, \mathcal{L}_{TD5}$ using Huber Loss}
\STATE{Set $\mathcal{L}_{TD}=\mathcal{L}_{TD1}+\mathcal{L}_{TD2}+\mathcal{L}_{TD3}+\mathcal{L}_{TD4}+\mathcal{L}_{TD5}$}
\STATE{Calculate margin loss $\mathcal{L}_{SLM1}, \mathcal{L}_{SLM2}, \mathcal{L}_{SLM3}, \mathcal{L}_{SLM4}, \mathcal{L}_{SLM5}$ for $q_1, q_2, q_3, q_4, q_5$ using Equation~\ref{eqn:slm}}
\STATE{Set $\mathcal{L}_{SLM}=\mathcal{L}_{SLM1}+\mathcal{L}_{SLM2}+\mathcal{L}_{SLM3}+\mathcal{L}_{SLM4}+\mathcal{L}_{SLM5}$}
\STATE{Set $\mathcal{L}=\mathcal{L}_{TD}+w\mathcal{L}_{SLM}$}
\STATE{Perform a gradient descent step to update $q_1, q_2, q_3, q_4, q_5$}
\STATE{Update target networks after certain steps}
\ENDFOR
\FOR{$N$ episodes}
\STATE{In state $s$, calculate greedy action $a_{xy}=a_1, a_\theta=a_2, a_z=a_3, a_\phi=a_4, a_\psi=a_5$:}
\STATE{$a_1=\argmax q_1(s)$}
\STATE{$a_2=\argmax q_2(e(s), H, f_2(s, a_1))$}
\STATE{$a_3=\argmax q_3(e(s), H, f_3(s, a_1, a_2))$}
\STATE{$a_4=\argmax q_4(e(s), H, f_4(s, a_1, a_2, a_3))$}
\STATE{$a_5=\argmax q_5(e(s), H, f_5(s, a_1, a_2, a_3, a_4))$}
\STATE{$e(s)$ is from the forward pass of ${q_1}(s)$}
\STATE{Play $a_{xy}, a_{\theta}, a_{z}, a_\phi, a_\psi$, observe $s', r$, save transition in $\mathcal{D}_s$}
\STATE{Sample half minibatch from $\mathcal{D}_e$ and half minibatch from $\mathcal{D}_s$}
\STATE{Calculate $\mathcal{L}_{TD}$ the same way for all transitions}
\STATE{Calculate $\mathcal{L}_{SLM}$ the same way only for transitions from $\mathcal{D}_e$}
\STATE{Set $\mathcal{L}=\mathcal{L}_{TD}+w\mathcal{L}_{SLM}$}
\STATE{Perform a gradient descent step to update $q_1, q_2, q_3, q_4, q_5$}
\STATE{Update target networks after certain steps}
\ENDFOR
\end{algorithmic}
\caption{ASRSE3 SDQfD}
\label{alg:asrse3_sdqfd}
\end{algorithm}

Except for Section~\ref{sec:exp_robot} where the testing is in the real world, all training and testing are in the PyBullet simulation~\cite{pybullet}. We train our models using PyTorch~\cite{pytorch} with the Adam~\cite{adam} optimizer with a learning rate of $0.5\times10^{-5}$ and weight decay of $10^{-5}$. All $Q$ function losses use the Huber loss~\cite{huberloss} and all behavior cloning losses use the cross entropy loss. The discount factor $\gamma$ is 0.9. The batch size is 32. The replay buffer has a size of 100,000 transitions. For ADET variations, the weight $w$ for cross-entropy loss is 0.01, and the softmax policy temperature $\beta$ is 10 for FCN methods and $Q_1$ in ASRSE3 methods. For SDQfD and DQfD variations, the weight $w$ for margin loss is 0.1, and $l$ is 0.1. For methods involving TD loss, the target $Q$ network updates every 100 training steps. The probability $\epsilon$ for taking a random action anneals from 0.5 to 0 in 20k episodes in vanilla DQN (without pretraining), and is constantly 0 for other methods.

When ASRSE3 is applied to the $(x, y, \theta)$ action space, $Q_3, Q_4, Q_5$ are skipped and $\bar{\mathcal{M}}$ transitions to $s'$ directly after taking action $a_2$ from state $(s, a_1)$. In that case, the TD target is calculated as: $y_1=y_2=r+\gamma\max_{a'_2}Q_2((s', a'^*_1), a'_2)$, where $a'^*_1=\argmax_{a'_1}Q_1(s', a'_1)$.

\section{Experimental Tasks}
\label{appen:task}

\begin{figure}[t]
    \centering
\includegraphics[width=0.7\linewidth]{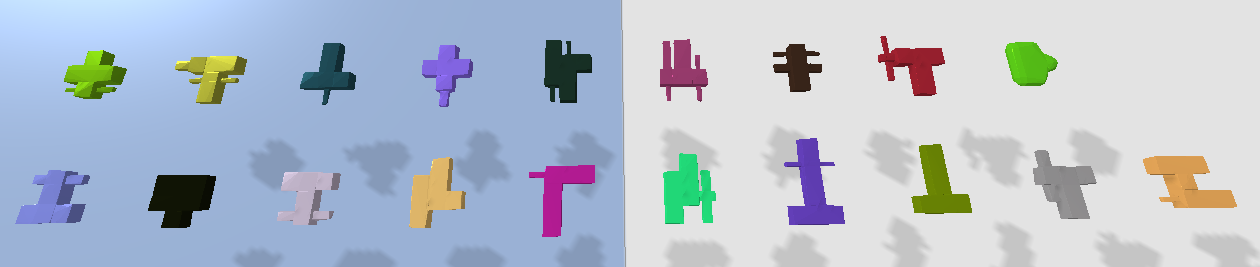}
\caption{The random shapes used in simulation}
\label{fig:envs_im_objset}
\end{figure}

The shapes of objects in the first seven environments in Figure~\ref{fig:envs} are fixed as shown in the figure. 

In ImDis, ImRan, ImH2, and ImH3, the agent needs to use shapes with randomly specified dimensions (Figure~\ref{fig:envs_im_objset}) in order to build the base structure below the triangle roof (we call these the ``improvize'' environments). Specifically, in ImDis and ImRan, there will be one fixed-shape roof and four random-shape objects. In ImDis, there are two discretized heights randomly assigned to each random shape: shorter (1.5cm) and higher (3cm). The robot can build a base by either stacking 2 shorter objects or using 1 higher object. Figure~\ref{fig:envs_imdis} shows an example of building such structure using one of each possible bases. In ImRan, the heights for those objects are randomly sampled: first, 2 objects will sample their heights in cm separately from a uniform distribution $U(1.5, 3)$, then the other two will have heights of 4.5cm minus each sampled heights. Thus each two of the four objects can form a 4.5cm-high base. ImH2 and ImH3 are respectively the improvised version of H2 and H3. Instead of using the cube to form the base, in ImH2 and ImH3 the agent needs to use the random shapes. The heights of the random shapes in those two domains are sampled from 3cm to 3.3cm. In addition, the size of the brick in ImH3 is also randomly sampled (7.5cm to 10.5cm in length, 2.5cm to 3.5cm in width, 2cm to 3.5cm in height).

In all environments, the relevant blocks are initially placed in the workspace with a random position and orientation. The maximum number of steps per episode is 20 for H4 and 10 for other environments. 




\section{SDQfD versus baselines in 5H1 and ImRan}
\label{appen:sdqfd_5h1_imran}

\begin{figure}[t]
\centering
\subfloat[FCN 5H1]{\includegraphics[width=0.2\textwidth]{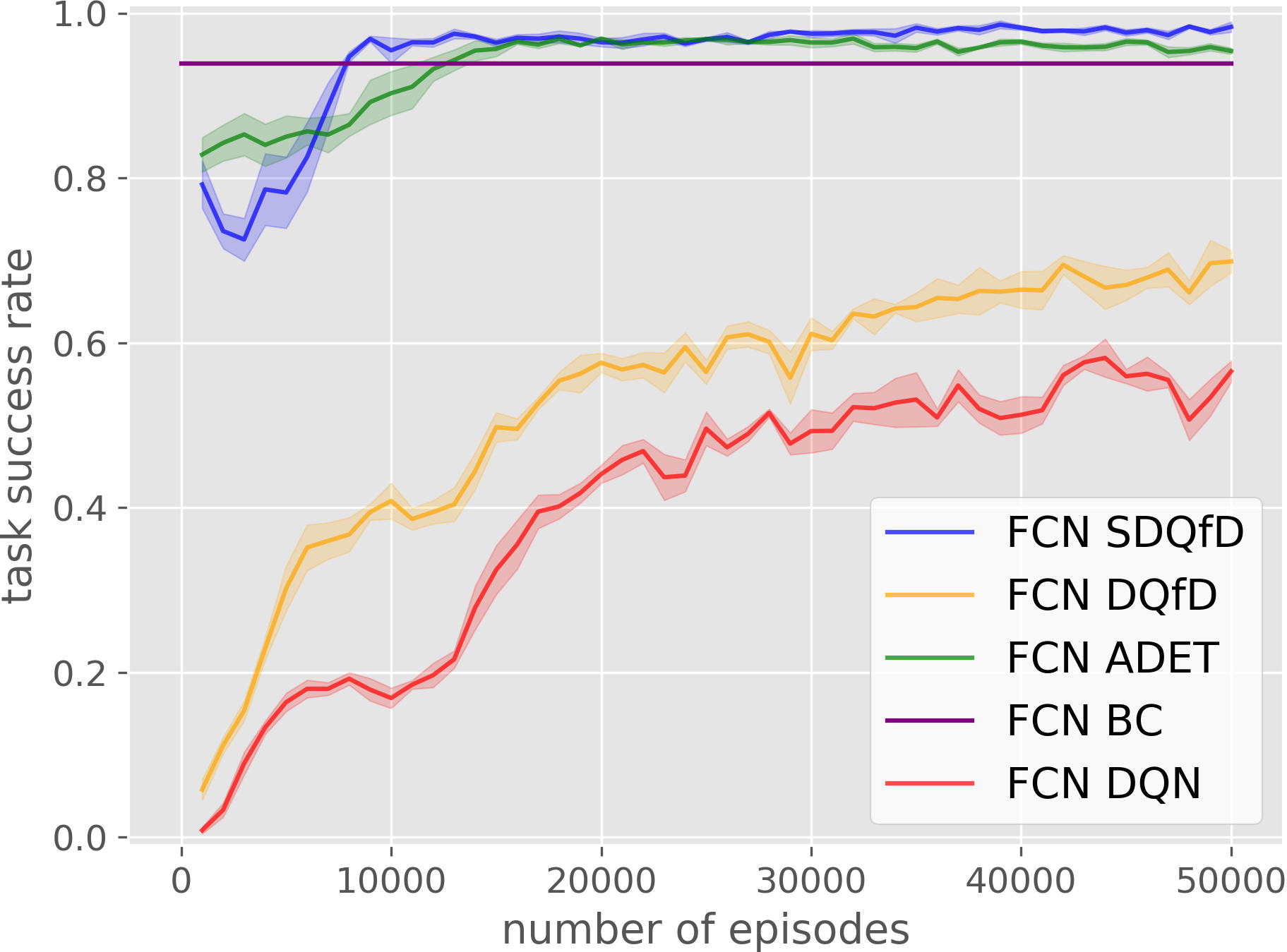}}
\subfloat[ASRSE3 5H1]{\includegraphics[width=0.2\textwidth]{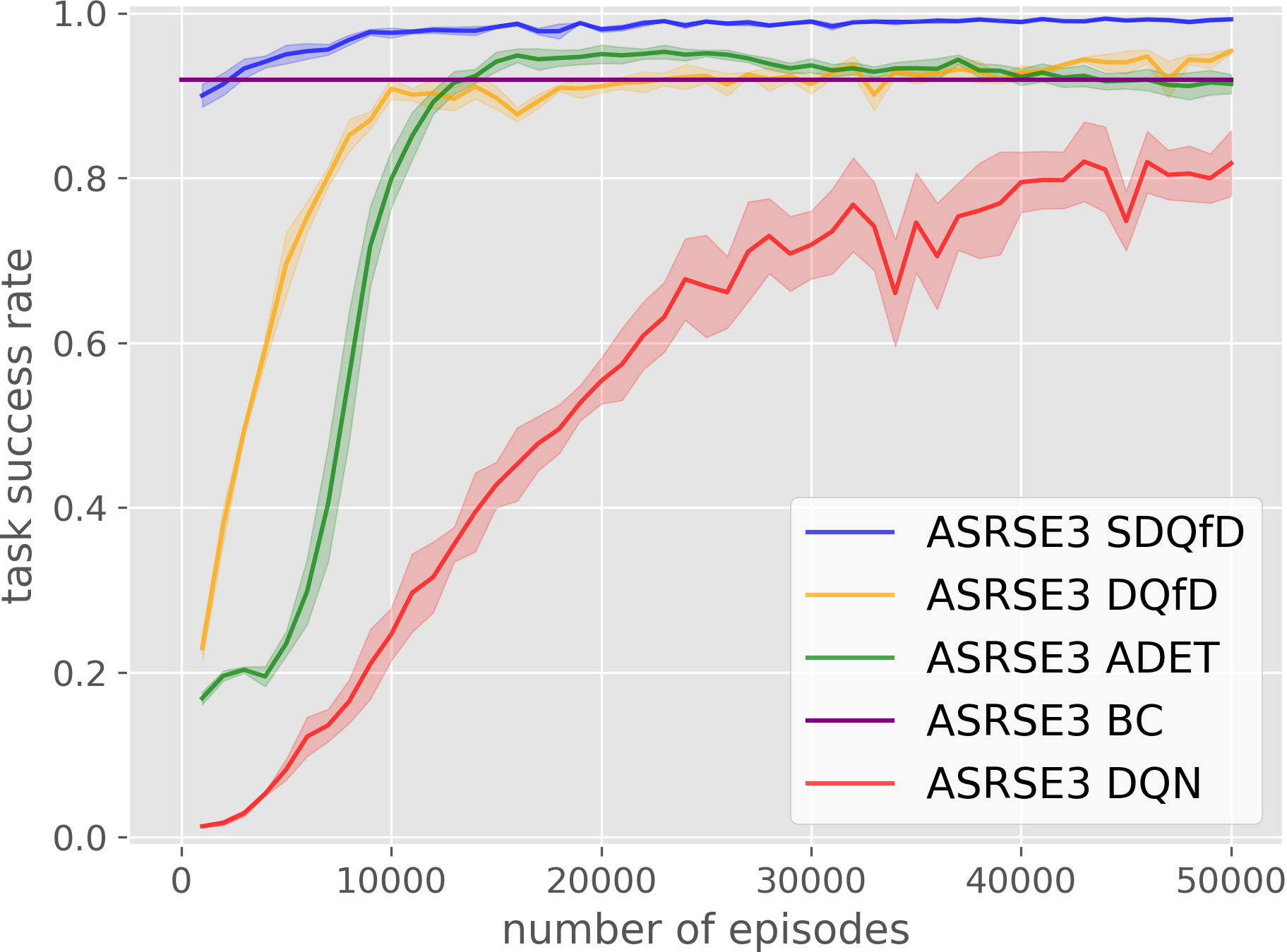}}
\subfloat[FCN ImRan]{\includegraphics[width=0.2\textwidth]{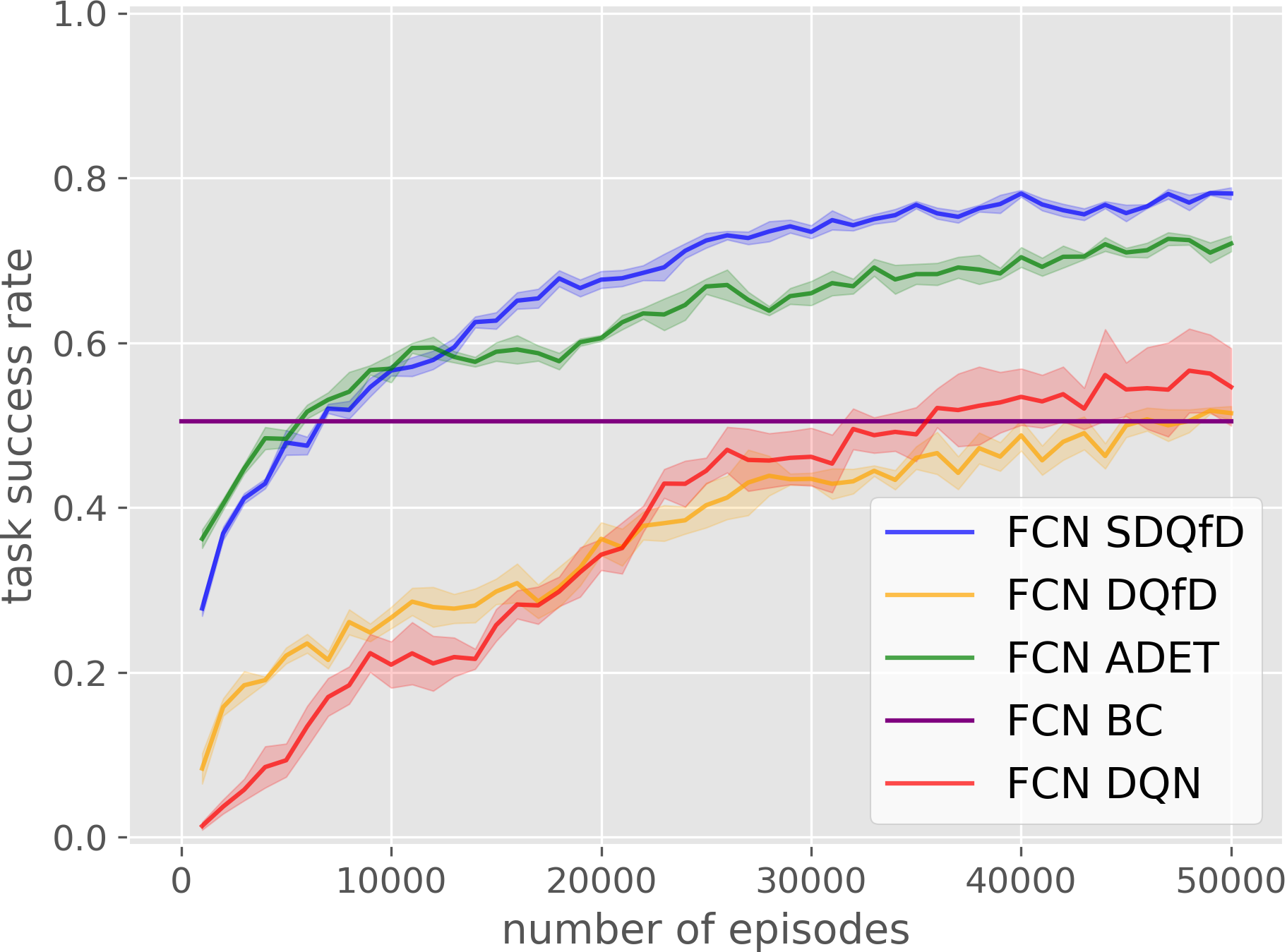}}
\subfloat[ASRSE3 ImRan]{\includegraphics[width=0.2\textwidth]{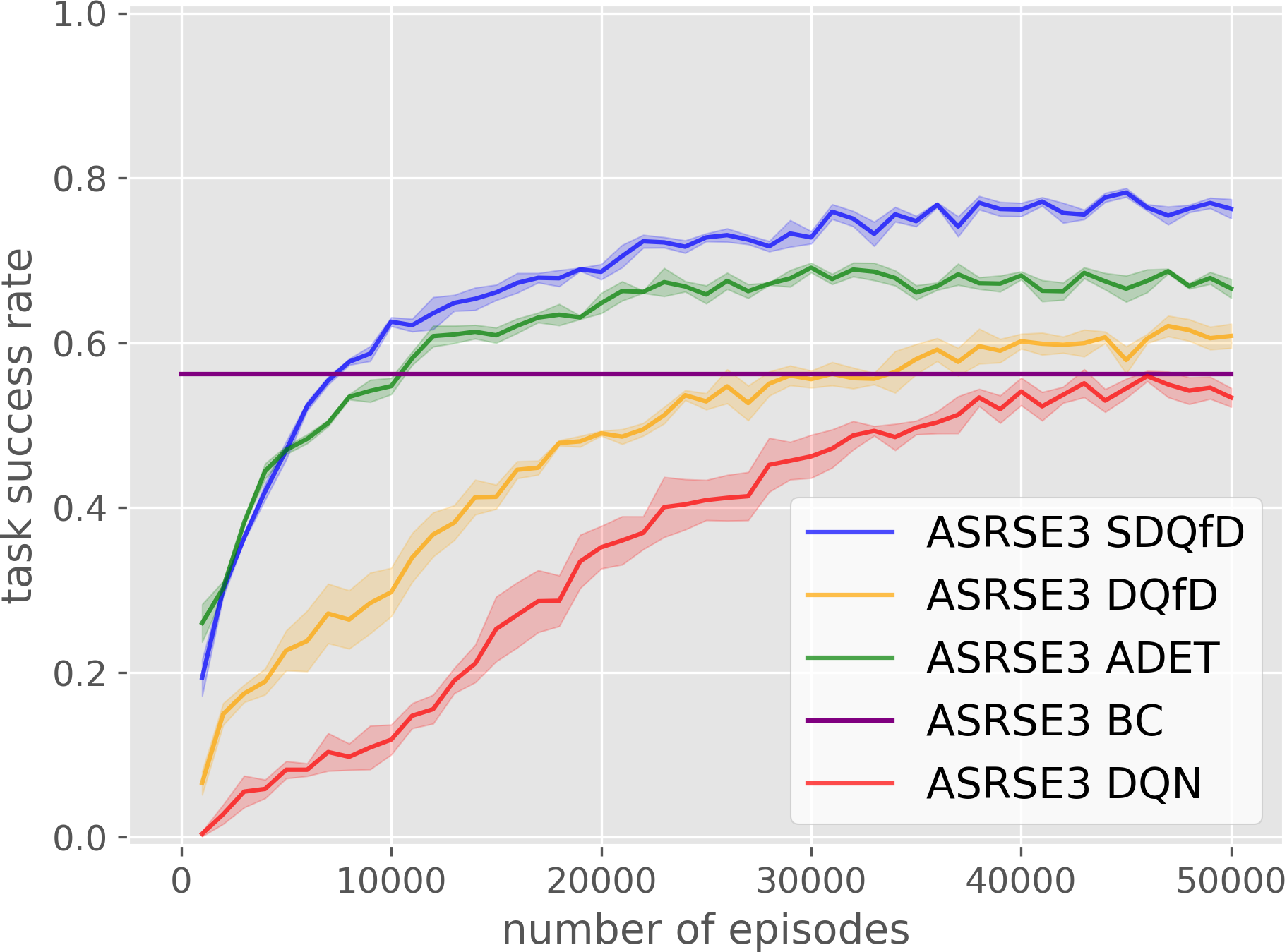}}
\caption{(a, c) FCN SDQfD versus with baseline methods for pixelwise $Q$ function encoding. (b, d) Same comparison with ASRSE3. All algorithms have a pretraining phase of 10k training steps on expert transitions. Results are averaged over 4 runs and plotted with a moving average of 1000 episodes.}
\label{fig:exp_sdqfd_5h1_imran}
\end{figure}

The results of the comparison of FCN SDQfD versus baselines and the comparison of ASRSE3 SDQfD versus baselines are shown in Figure~\ref{fig:exp_sdqfd_5h1_imran}. Similar to the results shown in Figure~\ref{fig:exp_sdqfd}, in the FCN version of all algorithms, SDQfD and ADET outperforms other methods significantly, while SDQfD outperforms ADET slightly. In ASRSE3, the performance of DQfD increases a lot, but ASRSE3 SDQfD is still the best-performing algorithm.
\end{document}